\documentclass{article} %
\usepackage{iclr2027_conference,times}
\newcommand{\figurefont}{\fontencoding{OT1}\fontfamily{phv}\selectfont}

\usepackage{amsmath,amsfonts,bm}

\def\eqref#1{equation~\ref{#1}}

\def\1{\bm{1}}

\def\rvp{{\mathbf{p}}}
\def\rvq{{\mathbf{q}}}

\DeclareMathAlphabet{\mathsfit}{\encodingdefault}{\sfdefault}{m}{sl}
\SetMathAlphabet{\mathsfit}{bold}{\encodingdefault}{\sfdefault}{bx}{n}

\usepackage{hyperref}
\hypersetup{colorlinks=true, linkcolor=blue!60!black, citecolor=blue!60!black, urlcolor=magenta!60!black}
\usepackage{cleveref}
\usepackage{url}
\usepackage{graphicx}
\usepackage{subcaption}
\usepackage{float}
\usepackage{booktabs}
\usepackage{multirow}
\usepackage{amssymb}
\usepackage{colortbl}
\usepackage{tikz}
\usepackage{tcolorbox}
\usepackage{pgfplots}
\pgfplotsset{compat=1.18}
\usetikzlibrary{arrows.meta,calc,positioning}
\tcbuselibrary{skins}
\definecolor{lightblue}{RGB}{218,238,247}
\definecolor{lightgreen}{RGB}{220,241,220}
\definecolor{lightpink}{RGB}{247,220,229}
\definecolor{lightgray}{RGB}{240,240,240}
\definecolor{lightyellow}{RGB}{255,244,204}
\definecolor{modellavender}{RGB}{232,225,250}
\definecolor{modelpeach}{RGB}{255,229,210}
\definecolor{modelaqua}{RGB}{210,242,236}
\definecolor{modelsand}{RGB}{245,235,205}
\newcommand{\githubicon}{\raisebox{-0.12em}{\tikz[x=0.0625em,y=-0.0625em]\fill (8.00,0.00) -- (5.87,0.29) -- (3.96,1.09) -- (2.34,2.34) -- (1.09,3.96) -- (0.29,5.87) -- (0.00,8.00) -- (0.18,9.72) -- (0.71,11.31) -- (1.54,12.73) -- (2.64,13.94) -- (3.96,14.91) -- (5.47,15.59) -- (5.65,15.60) -- (5.79,15.57) -- (5.89,15.51) -- (5.97,15.42) -- (6.01,15.32) -- (6.02,15.21) -- (6.02,15.08) -- (6.02,14.89) -- (6.01,14.64) -- (6.01,14.36) -- (6.01,14.05) -- (6.01,13.72) -- (5.12,13.81) -- (4.45,13.74) -- (3.97,13.56) -- (3.64,13.31) -- (3.44,13.03) -- (3.32,12.78) -- (3.25,12.63) -- (3.14,12.43) -- (3.00,12.20) -- (2.84,11.98) -- (2.67,11.78) -- (2.50,11.65) -- (2.36,11.56) -- (2.22,11.45) -- (2.14,11.33) -- (2.13,11.23) -- (2.24,11.15) -- (2.49,11.12) -- (2.79,11.16) -- (3.06,11.27) -- (3.29,11.44) -- (3.47,11.62) -- (3.62,11.80) -- (3.72,11.94) -- (4.11,12.42) -- (4.54,12.70) -- (4.98,12.80) -- (5.40,12.79) -- (5.77,12.71) -- (6.05,12.60) -- (6.10,12.35) -- (6.16,12.14) -- (6.25,11.95) -- (6.34,11.78) -- (6.45,11.64) -- (6.56,11.53) -- (5.67,11.38) -- (4.83,11.10) -- (4.07,10.63) -- (3.47,9.92) -- (3.07,8.92) -- (2.92,7.58) -- (2.95,7.16) -- (3.02,6.76) -- (3.14,6.39) -- (3.30,6.04) -- (3.50,5.72) -- (3.74,5.43) -- (3.69,5.28) -- (3.63,5.04) -- (3.58,4.71) -- (3.59,4.30) -- (3.66,3.84) -- (3.82,3.31) -- (3.88,3.30) -- (4.05,3.29) -- (4.35,3.33) -- (4.77,3.46) -- (5.33,3.71) -- (6.02,4.13) -- (6.34,4.05) -- (6.67,3.98) -- (7.01,3.93) -- (7.34,3.89) -- (7.68,3.87) -- (8.02,3.86) -- (8.36,3.87) -- (8.70,3.89) -- (9.03,3.93) -- (9.37,3.98) -- (9.70,4.05) -- (10.02,4.13) -- (10.71,3.71) -- (11.27,3.46) -- (11.69,3.33) -- (11.99,3.29) -- (12.16,3.30) -- (12.22,3.31) -- (12.38,3.84) -- (12.45,4.30) -- (12.46,4.71) -- (12.41,5.04) -- (12.35,5.28) -- (12.30,5.43) -- (12.54,5.72) -- (12.74,6.04) -- (12.90,6.38) -- (13.02,6.76) -- (13.09,7.15) -- (13.12,7.58) -- (12.97,8.92) -- (12.57,9.92) -- (11.96,10.63) -- (11.21,11.10) -- (10.36,11.38) -- (9.47,11.53) -- (9.61,11.67) -- (9.74,11.86) -- (9.85,12.08) -- (9.93,12.35) -- (9.99,12.66) -- (10.01,13.01) -- (10.01,13.53) -- (10.01,14.00) -- (10.00,14.41) -- (10.00,14.76) -- (10.00,15.03) -- (10.00,15.21) -- (10.01,15.32) -- (10.05,15.42) -- (10.12,15.51) -- (10.23,15.57) -- (10.37,15.60) -- (10.55,15.59) -- (10.94,15.45) -- (11.32,15.28) -- (11.70,15.10) -- (12.06,14.90) -- (12.41,14.68) -- (12.75,14.44) -- (13.08,14.18) -- (13.39,13.91) -- (13.69,13.62) -- (13.98,13.32) -- (14.24,13.00) -- (14.50,12.67) -- (14.73,12.33) -- (14.94,11.97) -- (15.14,11.61) -- (15.32,11.23) -- (15.48,10.85) -- (15.61,10.45) -- (15.73,10.06) -- (15.83,9.65) -- (15.90,9.24) -- (15.96,8.83) -- (15.99,8.42) -- (16.00,8.00) -- (15.71,5.87) -- (14.91,3.96) -- (13.66,2.34) -- (12.04,1.09) -- (10.13,0.29) -- (8.00,0.00) -- cycle;}}

\title{Flat Features, Detailed Images?
Simple Detail Control for Diffusion Transformers}

\definecolor{alphaColor}{HTML}{FF9100}
\definecolor{betaColor}{HTML}{00D5FF}
\definecolor{gammaColor}{HTML}{00B050}
\newcommand{\instalpha}{{\color{alphaColor}{\alpha}}}
\newcommand{\instbeta}{{\color{betaColor}{\beta}}}
\newcommand{\instgamma}{{\color{gammaColor}{\gamma}}}

\author{\parbox[t]{\dimexpr\textwidth-2\tabcolsep\relax}{\raggedright
  Qingtao~Yu$^{\instalpha}$ \quad
  Zhen~Wang$^{\instalpha}$ \quad
  Jiyang~Zheng$^{\instbeta}$ \quad
  Minghao~Sun$^{\instalpha\instgamma}$ \\
  Hongdong~Li$^{\instalpha}$ \quad
  Dylan~Campbell$^{\instalpha}$ \\[2pt]
  \normalfont
  $^{\instalpha}$The Australian National University \quad
  $^{\instbeta}$Microsoft \quad
  $^{\instgamma}$The University of Queensland \\
  {\fontsize{6.5}{8}\selectfont\texttt{\{terry.yu,\allowbreak zhen.wang,\allowbreak minghao.sun,\allowbreak hongdong.li,\allowbreak dylan.campbell\}@anu.edu.au,\allowbreak\ jiyang.zheng@microsoft.com}}}
}

\iclrfinalcopy %
\begin{document}

\maketitle
\lhead{Preprint}
\vspace{-8mm} %

\input{figures/fig_teaser.tex}

\begin{abstract}

Prompt engineering is commonly used to encourage text-to-image models to generate more detailed images.
However, it does not provide reliable, continuous, or disentangled control.
We analyze feature tokens produced with and without detail-enhancing prompts across six diffusion transformers and observe a common pattern: these prompts tend to
increase the entropy of each token's energy distribution
while largely preserving its ordering.
That is, flattening the distribution increases the level of detail.
Motivated by this counter-intuitive finding, we introduce a simple training-free approach that continuously adjusts the level of detail according to a scalar.
Specifically, it shifts the squared channel magnitudes of each token toward a higher entropy distribution with a time-aware scheduler to concentrate the intervention at intermediate sampling steps.
Across extensive experiments on different types of diffusion transformers, we show that our approach provides continuous control over detail metrics while largely preserving image quality and semantic structure.
It requires only a few lines of code and adds negligible inference time overhead.
Code is available at \href{https://github.com/TerrysLearning/TokenFlat}{\githubicon\,\texttt{TerrysLearning/TokenFlat}}.
\end{abstract}

\section{Introduction}

Prompt engineering is a straightforward approach that can be used to enhance visual detail in images generated by text-to-image (T2I) models \citep{promptdesign,promptopt}, 
for example, by adding phrases such as ``highly detailed'' or ``fine texture''.
However, such instructions provide vague control, making their strength difficult to specify precisely. 
Their effects are also unpredictable and model-dependent, as different models may interpret the same phrase differently. 
Moreover, prompt-based control is not localized and may alter the global semantic structure, 
complicating iterative refinement.
We aim to understand how such detail-focused prompt engineering impacts feature tokens internally, 
and whether universal patterns exist that can be leveraged for finer control.

Flow-matching models \citep{flow_gen} with diffusion transformers (DiTs) \citep{dit} have emerged as a dominant paradigm for T2I generation.
We first investigate how such prompts affect feature tokens in pretrained DiTs. 
Given the same image embedding, we compare the features obtained with and without detail instructions in the prompt. 
Across six variants of DiTs, we analyze all tokens at every layer and inference timestep. 
We find that detail-based instructions consistently increase the entropy of the token's energy distribution, that is, the distribution of its squared channel amplitudes, while largely preserving their rank ordering. 
In contrast, adding content descriptions rather than detail descriptions exhibits this pattern weakly or not at all.
This observation suggests that detail richness could be controlled in a training-free manner by directly rescaling channel magnitudes.
It may seem counter-intuitive that flattening a token's feature distribution should make the associated image patch \textit{less} flat. 
We speculate that activating a sparse set of feature dimensions encodes a specific texture, so activating a larger set (i.e., a higher entropy distribution) leads to a mixture of textures.

Motivated by this finding, we introduce
T$\flat$ (Token Flattening), a simple mechanism for controlling the level of detail in the generated images
according to a user-specified scalar $\alpha$. 
T$\flat$ linearly interpolates the per-token distribution of squared channel amplitudes (i.e., its energy distribution) with the uniform distribution, increasing its entropy.
To improve stability, we design a time-aware scheduler that emphasizes the intervention at intermediate sampling stages. 
T$\flat$ requires only a few lines of code and can be applied directly to DiTs as a plug-in.

We evaluate T$\flat$ on three representative diffusion transformers: 
FLUX.2 Klein 9B, FLUX.2 Klein 9B Base, and Qwen-Image \citep{flux2klein, qwenimage},
covering distilled and undistilled variants with both single and double stream architectures. 
Varying $\alpha$ produces
continuous
changes across all evaluated detail metrics 
while largely preserving text alignment, perceptual quality, and semantic structure.
Some qualitative examples are demonstrated in \cref{fig:teaser}.
Compared with existing training-free detail-enhancement methods, 
T$\flat$ achieves greater structure-preserving detail improvements and increased image quality,
with negligible additional inference time overhead.

\section{Related Work}
\paragraph{Diffusion transformers.}
Previously, diffusion models were based on U-Net architectures \citep{sd,sdxl,playgroundv25}.
Nowadays, Diffusion Transformers (DiTs) \citep{dit} are dominant due to their better performance and scalability.
Text-conditioned DiTs have evolved into several architectural variants.
For example, text embeddings can be incorporated into the image stream via cross-attention~\citep{pixelart}.
In MMDiT architectures, text and image embeddings are processed in two separate streams,
with joint attention enabling interaction between the two modalities~\citep{sd3,qwenimage}.
In single-stream architectures, text and image embeddings are concatenated into a unified sequence and processed jointly through self-attention~\citep{zimage}.
Some models combine both designs, using dual-stream blocks followed by single-stream blocks~\citep{flux,flux2klein}.
Furthermore, some alternatives leverage linear attention instead to speed up inference \citep{sana}.

\paragraph{Manipulating T2I models at inference time.}
Given a pretrained text-to-image model,
its inference process can be manipulated to achieve various desired effects.
For example, some methods scale up test-time computation through search or optimization
to obtain more desirable generations according to a preference model~\citep{das,ttsma,fk}.
Other methods improve generation quality and mitigate sampling artifacts by constructing degraded internal predictions as negative references~\citep{sag,pag,ssg,seg}.
\citet{resdit, hiflow} enable models to generate images beyond their training resolution.
Prior work has also adapted T2I models for image editing \citep{p2p}, inpainting \citep{repaint}, super-resolution \citep{sr3},
interpolation \citep{geodesicdiffusion}, 3D generation \citep{dreamfusion}, and personalized generation \citep{dreambooth}.

Our work aims to provide a knob for controlling the detail level of generated images.
Some closely related methods enhance the detail of images by amplifying high-frequency energy,
including FreeU \citep{freeu}, Fresca \citep{fresca}, and FDG \citep{fdg}.
However, FreeU is designed for U-Nets, while Fresca and FDG target undistilled models that use CFG \citep{cfg}.
Furthermore, \citet{dg} enhances detail through negative guidance constructed by masking the massively activated channels,
but requires an additional model forward pass, which causes large extra inference computation.
\citet{detailpp,promptenhancer} do this through prompt manipulation
but do not provide continuous control.
In contrast, our method generalizes across DiT architectures while being efficient and explicitly controllable.

\section{The Effect of Detail-Enhancing Prompts}
\label{sec:analysis}
\begin{figure}[t]
    \centering
    \resizebox{\textwidth}{!}{%
\definecolor{tensorfront}{RGB}{219,232,251}
\definecolor{tensortop}{RGB}{238,244,255}
\definecolor{tensorside}{RGB}{207,222,248}
\definecolor{latentfront}{RGB}{232,224,248}
\definecolor{latenttop}{RGB}{244,240,253}
\definecolor{latentside}{RGB}{215,201,239}
\definecolor{promptgreen}{RGB}{228,243,231}
\definecolor{tokenyellow}{RGB}{244,211,94}
\definecolor{noteyellow}{RGB}{255,243,196}
\definecolor{distributionblue}{RGB}{35,85,155}
\definecolor{distributionred}{RGB}{165,55,65}

\begin{tikzpicture}[
  x=0.68cm,
  y=0.68cm,
  >=Stealth,
  font=\figurefont\small,
  flow/.style={draw, line width=0.8pt, -{Stealth[length=2.2mm]}},
  gridline/.style={draw=black!85, line width=0.35pt},
  prompt/.style={fill=promptgreen, draw=black!65, line width=1pt,
    rounded corners=2mm, inner sep=2mm},
]
  \begin{scope}[xshift=-0.5cm, yshift=0.36cm, scale around={0.75:(1.8,3.7)}]
  \filldraw[fill=latenttop] (0.4,4.8) -- (0.8,5.0) -- (3.2,5.0) -- (2.8,4.8) -- cycle;
  \filldraw[fill=latentside] (2.8,2.4) -- (3.2,2.6) -- (3.2,5.0) -- (2.8,4.8) -- cycle;
  \filldraw[fill=latentfront] (0.4,2.4) rectangle (2.8,4.8);
  \foreach \i in {1,...,4} {
    \draw[gridline] ({0.4+0.48*\i},2.4) -- ({0.4+0.48*\i},4.8);
    \draw[gridline] (0.4,{2.4+0.48*\i}) -- (2.8,{2.4+0.48*\i});
    \draw[gridline] ({0.4+0.48*\i},4.8) -- ({0.8+0.48*\i},5.0);
    \draw[gridline] (2.8,{2.4+0.48*\i}) -- (3.2,{2.6+0.48*\i});
  }
  \end{scope}
  \node[below, align=center] at (1.06,2.95) {Image Latent\\(VAE Space)};

  \foreach \rowy/\name in {5.33/ditup,2.88/ditdown} {
    \node[draw=black!65, fill=black!7, line width=1pt, rounded corners=2mm,
      minimum width=1.24cm, minimum height=1.3cm,
      align=center] (\name) at (11.56,\rowy) {DiT\\Block};
  }

  \node[prompt, anchor=west, minimum width=4.37cm] (promptup) at (3.2,5.55) {``A dog.''};
  \node[prompt, anchor=west, align=left, font=\figurefont\footnotesize] (promptdown) at (3.2,2.36)
    {``A dog. \textit{Ultra detailed, hyper}\\
     \textit{realistic, high resolution.}''};
  \draw[draw=black!65, line width=1pt, -{Stealth[length=2.2mm]}]
    (promptup.east) -- (promptup.east -| ditup.west);
  \draw[draw=black!65, line width=1pt, -{Stealth[length=2.2mm]}]
    (promptdown.east) -- (promptdown.east -| ditdown.west);

  \draw[draw=black!65, line width=1pt] (2.55,4.1) -- (9.7,4.1);
  \draw[draw=black!65, line width=1pt] (9.7,4.8) -- (9.7,3.4);
  \draw[draw=black!65, line width=1pt, -{Stealth[length=2.2mm]}]
    (9.7,4.8) -- (9.7,4.8 -| ditup.west);
  \draw[draw=black!65, line width=1pt, -{Stealth[length=2.2mm]}]
    (9.7,3.4) -- (9.7,3.4 -| ditdown.west);

  \draw[draw=black!65, line width=1pt, -{Stealth[length=2.2mm]}] (12.48,5.33) -- (13.55,5.33);
  \draw[draw=black!65, line width=1pt, -{Stealth[length=2.2mm]}] (12.48,2.88) -- (13.55,2.88);

  \begin{scope}[xshift=3.5cm]
  \foreach \basey/\tokenfill in {4.2/distributionblue,1.5/distributionred} {
    \begin{scope}[scale around={0.75:(10.45,{\basey+1.525})}]
    \filldraw[fill=tensortop] (8.4,{\basey+2.4}) -- (10.1,{\basey+3.05}) --
      (12.5,{\basey+3.05}) -- (10.8,{\basey+2.4}) -- cycle;
    \filldraw[fill=tensorside] (10.8,\basey) -- (12.5,{\basey+0.65}) --
      (12.5,{\basey+3.05}) -- (10.8,{\basey+2.4}) -- cycle;
    \filldraw[fill=tensorfront] (8.4,\basey) rectangle (10.8,{\basey+2.4});
    \foreach \i in {1,...,4} {
      \draw[gridline] ({8.4+0.48*\i},\basey) -- ({8.4+0.48*\i},{\basey+2.4});
      \draw[gridline] (8.4,{\basey+0.48*\i}) -- (10.8,{\basey+0.48*\i});
      \draw[gridline] ({8.4+0.48*\i},{\basey+2.4}) -- ({10.1+0.48*\i},{\basey+3.05});
      \draw[gridline] (10.8,{\basey+0.48*\i}) -- (12.5,{\basey+0.65+0.48*\i});
    }
    \filldraw[fill=\tokenfill!75] (10.32,{\basey+0.48}) rectangle (10.8,{\basey+0.96});
    \filldraw[fill=\tokenfill!65!black] (10.8,{\basey+0.48}) --
      (12.5,{\basey+1.13}) -- (12.5,{\basey+1.61}) -- (10.8,{\basey+0.96}) -- cycle;
    \end{scope}
  }
  \node[below] at (10.45,1.60) {Feature Output};

  \node[align=center, anchor=south] at (15.45,6.05) {Normalised Channel\\Square Magnitude};
  \draw[flow, distributionblue] (12.05,5.40) -- (13.10,5.40);
  \begin{scope}[yshift=-0.936cm]
  \draw[gray] (13.35,6.00) -- (17.85,6.00);
  \foreach \i/\h in {0/1.26,1/0.87,2/0.60,3/0.41,4/0.29,5/0.21,6/0.16,7/0.11,8/0.07,9/0.04} {
    \filldraw[fill=distributionblue!75, draw=distributionblue]
      ({13.50 + 0.42*\i},6.00) rectangle ({13.78 + 0.42*\i},{6.00 + \h});
  }
  \end{scope}

  \begin{scope}[yshift=0.664cm]
  \draw[flow, distributionred] (12.05,1.74) -- (13.10,1.74);
  \draw[gray] (13.35,1.35) -- (17.85,1.35);
  \foreach \i/\h in {0/0.7997,1/0.7372,2/0.6747,3/0.5997,4/0.5247,5/0.4623,6/0.3249,7/0.2499,8/0.1749,9/0.0750} {
    \filldraw[fill=distributionred!75, draw=distributionred]
      ({13.50 + 0.42*\i},1.35) rectangle ({13.78 + 0.42*\i},{1.35 + \h});
  }
  \node[fill=noteyellow, font=\figurefont\footnotesize\itshape, inner sep=1.5mm]
    at (15.40,0.4) {Entropy $\uparrow$};
  \end{scope}
  \end{scope}
\end{tikzpicture}%
    }
    \vspace{-5.5mm}
    \caption{Overview of the analysis experiment described in \cref{sec:analysis}.
    Given the same image latent, we compare DiT features obtained using an original prompt and its detail-guided version.
    For corresponding token pairs, detail instructions produce higher entropy over squared channel magnitudes.}
    \label{fig:observation}
\end{figure}
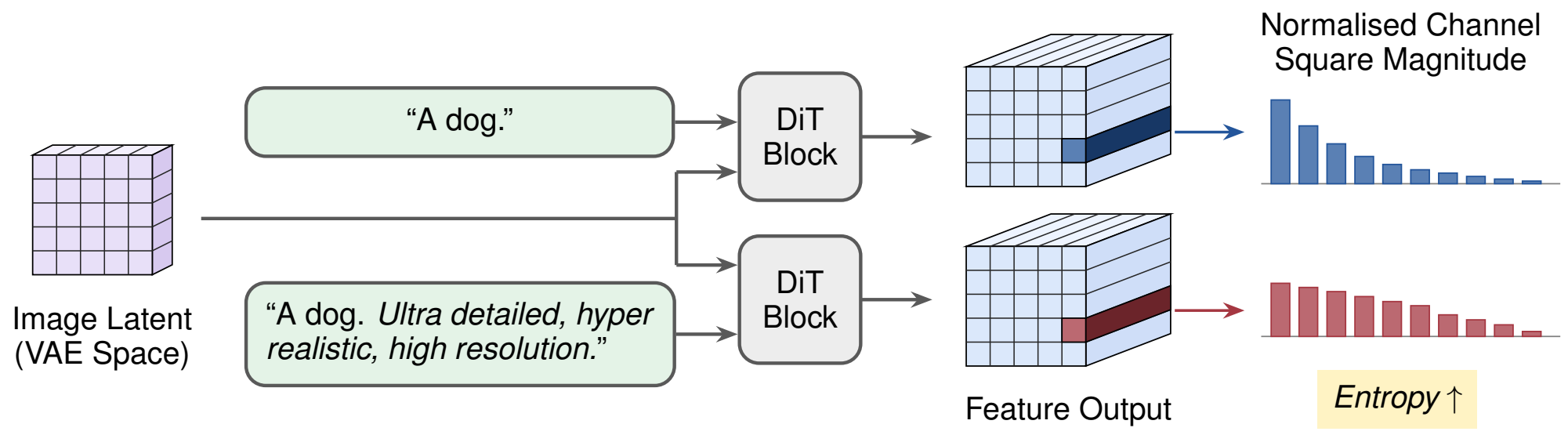

We first analyze feature tokens in pretrained DiTs under additional detail-oriented text conditioning.
Given an image VAE latent with $H \times W$ patches,
a DiT first up-projects it into a hidden feature of shape $(H \times W) \times D$,
where $D$ is the token dimension. 
For the same input image latent, we extract the output features after each transformer block
under text conditions with and without detail-enhancing instructions as suffix and compare their differences.
This process is illustrated in \cref{fig:observation}.
To obtain appropriate detail instructions,
we first perform a voting-based search for each model to obtain phrases that produce more detail, 
as measured by the Fourier high-frequency band energy ratio.
The searching process is described in \cref{app:phrase-search}.
We evaluate a total of 1,000 prompts from the COCO2017 validation set \citep{coco} across six commonly used open-weight undistilled and distilled DiT models \citep{flux2klein,sd3,zimage}.
We use $\rvq$ and $\hat{\rvq} \in \mathbb{R}^D$ to denote the produced feature token without and with detail instructions, respectively.
The main comparison results are summarized in \cref{tab:feature-statistics} and discussed further in \cref{app:more_obv}.

\begin{table*}[t]
    \centering
    \resizebox{0.9\textwidth}{!}{%
        \renewcommand{\arraystretch}{1.0}
        \begin{tabular}{llccccc}
            \toprule
            Model & $\overline{H}(\rvp)$ & Added text & $\overline{H}(\hat{\rvp})$ & $A_{\mathrm{all}}$ & $A_{\mathrm{prompt}}$ & Spearman \\
            \midrule
            \multirow{2}{*}{FLUX.2 Klein 9B Base} & \multirow{2}{*}{4.463} & content & 4.460 & 46.0\% & 45.6\% & 0.956 \\
            & & \cellcolor{lightblue}detail & \cellcolor{lightblue}\textbf{4.530} & \cellcolor{lightblue}\textbf{78.0\%} & \cellcolor{lightblue}\textbf{100.0\%} & \cellcolor{lightblue}\textbf{0.965} \\
            \midrule
            \multirow{2}{*}{FLUX.2 Klein 9B} & \multirow{2}{*}{4.307} & content & 4.292 & 39.7\% & 37.6\% & 0.799 \\
            & & \cellcolor{lightblue}detail & \cellcolor{lightblue}\textbf{4.433} & \cellcolor{lightblue}\textbf{83.7\%} & \cellcolor{lightblue}\textbf{100.0\%} & \cellcolor{lightblue}\textbf{0.878} \\
            \midrule
            \multirow{2}{*}{SD3.5 Large} & \multirow{2}{*}{3.065} & content & 3.070 & 59.2\% & 66.6\% & 0.929 \\
            & & \cellcolor{lightblue}detail & \cellcolor{lightblue}\textbf{3.080} & \cellcolor{lightblue}\textbf{71.4\%} & \cellcolor{lightblue}\textbf{98.2\%} & \cellcolor{lightblue}\textbf{0.966} \\
            \midrule
            \multirow{2}{*}{SD3.5 Large Turbo} & \multirow{2}{*}{3.006} & content & 3.013 & 57.7\% & 66.4\% & 0.881 \\
            & & \cellcolor{lightblue}detail & \cellcolor{lightblue}\textbf{3.027} & \cellcolor{lightblue}\textbf{70.0\%} & \cellcolor{lightblue}\textbf{96.2\%} & \cellcolor{lightblue}\textbf{0.928} \\
            \midrule
            \multirow{2}{*}{Z-Image Base} & \multirow{2}{*}{0.057} & content & 0.057 & 50.8\% & 55.8\% & 0.859 \\
            & & \cellcolor{lightblue}detail & \cellcolor{lightblue}{0.057} & \cellcolor{lightblue}{46.6\%} & \cellcolor{lightblue}\textbf{70.8\%} & \cellcolor{lightblue}\textbf{0.895} \\
            \midrule
            \multirow{2}{*}{Z-Image Turbo} & \multirow{2}{*}{1.087} & content & 1.088 & 59.0\% & 58.0\% & 0.868 \\
            & & \cellcolor{lightblue}detail & \cellcolor{lightblue}\textbf{1.096} & \cellcolor{lightblue}\textbf{62.2\%} & \cellcolor{lightblue}\textbf{94.4\%} & \cellcolor{lightblue}\textbf{0.903} \\
            \bottomrule
        \end{tabular}%
    }
    \vspace{-1mm}
    \caption{For each model, we first report the baseline mean entropy ($\overline{H}(\rvp)$) of the token density defined in \cref{eq:analysis-density}.
    Under each type of added text condition, we report the mean entropy ($\overline{H}(\hat{\rvp})$),
    together with pairwise statistics computed between spatially corresponding tokens of baseline:
    the entropy increasing agreement ratios (\cref{eq:entropy-agreement-all,eq:entropy-agreement-prompt}) and the mean Spearman correlation.
    Bold values indicate the patterns observed under detail instructions as discussed in \cref{sec:analysis}.}
    \label{tab:feature-statistics}
    \vspace{-2mm}
\end{table*}

\paragraph{Detail instructions tend to increase the entropy of token energy distributions.}
We compute the normalized energy distribution of the token using the squared amplitudes,
analogous to the Born rule from quantum physics \citep{bornrule}:
\begin{equation}
    \rvp_{i} = \frac{\rvq_{i}^2}{\|\rvq\|^2},
\label{eq:analysis-density}
\end{equation}
where $i = 1, \dots, D$ is the channel index.
Denoting $\rvp$ and $\hat{\rvp}$ as the densities computed from the corresponding token pair $\rvq$ and $\hat{\rvq}$,
we compare their entropies $H(\rvp)$ and $H(\hat{\rvp})$, averaging over all tokens, layers, timesteps, and prompts.
We found that detail-enhancing instructions mildly increase the mean entropy for five of the six models.
To provide stronger statistical evidence, we quantify the agreement ratio of this entropy increase.
Let $L$, $S$, and $K$ denote the numbers of transformer layers, inference steps, and prompts.
We calculate two aggregated pairwise agreements: 
\begin{align}
    A_{\mathrm{all}}
    &= \frac{1}{N}
    \sum_{n=1}^{N}
    1\left[
    H(\hat{\rvp}^{(n)})
    > H(\rvp^{(n)}) 
    \right], 
    \; N=HWLSK, 
    \label{eq:entropy-agreement-all} \\
    \text{and} \; A_{\mathrm{prompt}}
    &= \frac{1}{K}\sum_{k=1}^{K}
    1\left[
        \sum_{n=1}^{N}
        (
            H(\hat{\rvp}^{(n,k)})
            - H(\rvp^{(n,k)})
        )> 0
    \right], \; N = HWLS,
    \label{eq:entropy-agreement-prompt}
\end{align}
$A_{\mathrm{all}}$ measures the fraction of token pairs where $\hat{\rvp}$ has higher entropy than $\rvp$, 
which substantially exceeds 50\% for five out of six models.
At the prompt level, $A_{\mathrm{prompt}}$ measures the fraction of prompts where $\hat{\rvp}$ has higher summed entropy.
This agreement exceeds 94\% for five out of six models, which shows that the trend is consistent across prompts.
Therefore, we see that flatter (higher entropy) per-token energy distributions correspond to richer image details.
This might seem counter-intuitive, since a flatter distribution of RGB values within a patch would imply less variation and fewer details.
Nonetheless, we find that a flatter distribution across channels decodes to a less uniform RGB distribution across pixels.
We speculate that activating a sparse set of feature dimensions encodes a specific texture, while activating a larger set leads to a mixture of textures.

\paragraph{Detail instructions largely preserve the rank of channel magnitudes.}
For each pair of corresponding tokens, we measure the Spearman correlation between their channel-magnitude ranks and average it over all token pairs.
Under detail instructions, the correlation ranges from 0.878 to 0.966 across models, where 1.0 indicates perfect rank preservation and 0.0 indicates no correlation.
This result suggests that detail instructions tend not to alter the relative importance of different channels, 
which provides a useful cue to allow a simple training-free control.

\paragraph{These findings are specific to detail instructions.}
Does this effect arise simply from longer prompts or specifically from detail-oriented instructions?
To distinguish these factors, we add a content-adding baseline generated by an LLM that appends additional semantic information to the original prompt, such as adjectives and overall style descriptions.
Under additional content, neither the entropy change nor the agreement ratio shows a consistent trend.
Its Spearman correlation is also lower than that of detail conditioning, suggesting that extra semantic content changes the channel scale ordering more strongly.
Thus, longer prompts alone do not explain the consistent effects of adding detail instructions.
We summarize all the findings in this section as follows:

\begin{tcolorbox}[
    colback=gray!8,
    colframe=teal!60!black,
    boxrule=0.8pt,
    arc=4mm,
    left=2mm,
    right=2mm,
    top=2mm,
    bottom=2mm,
    drop shadow
]
\textbf{Summary:}
\textit{Detail instructions tend to increase the entropy of squared channel magnitudes within each feature token while largely preserving their rank.}
\end{tcolorbox}

\begin{figure}[t]
    \centering
    \resizebox{0.95\textwidth}{!}{%
\definecolor{sdcfill}{RGB}{252,240,3}
\definecolor{pseudocodefill}{RGB}{250,246,235}
\begin{tikzpicture}[
    x=1cm,
    y=1cm,
    font=\figurefont\footnotesize,
    line/.style={draw=black!65, line width=1pt},
    arrow/.style={line, -{Stealth[length=2mm, width=1.4mm]}},
]
    \begin{scope}[xshift=-0.7cm]
    \draw[line, fill=black!7, rounded corners=2mm]
        (0.30,0.55) rectangle (3.40,3.85);
    \node[anchor=east, font=\figurefont\normalsize] at (0.20,2.275) {$N\times$};

    \begin{scope}[yshift=-0.5cm]
    \node[anchor=north west, font=\figurefont\fontsize{8}{9.5}\selectfont] at (0.4,4.82) {DiT Block};

    \draw[line] (1.92,4.70) -- (1.92,3.30);
    \draw[line, fill=sdcfill, rounded corners=2mm]
        (0.98,3.46) rectangle (2.90,4.02);
    \node[font=\figurefont\fontsize{7.5}{9}\selectfont] at (1.94,3.74) {T$\flat$};
    \draw[line] (2.90,3.74) -- (4.15,3.74);

    \draw[line, fill=lightgreen, rounded corners=2mm]
        (0.98,1.70) rectangle (2.90,3.30);
    \node[align=center, font=\figurefont\fontsize{7.5}{9}\selectfont] at (1.94,2.44) {Layernorm\\Scale, Shift\\Attention\\$\ldots$};
    \draw[line] (1.92,1.70) -- (1.92,1.55);

    \draw[arrow, rounded corners=0.5mm]
        (1.92,4.15) -- (0.62,4.15) -- (0.62,1.35) -- (1.72,1.35);
    \draw[line, fill=black!7] (1.92,1.35) circle (0.20);
    \draw[line] (1.72,1.35) -- (2.12,1.35);
    \draw[line] (1.92,1.15) -- (1.92,1.55);
    \draw[line] (1.92,1.15) -- (1.92,0.80);
    \end{scope}
    \end{scope}

    \draw[line, fill=pseudocodefill, rounded corners=2mm]
        (3.42,0.40) rectangle (13.4,4.35);
    \node[anchor=north west, align=left, font=\figurefont\fontsize{8}{9.5}\selectfont] at (3.53,4.28) {T$\flat$ Pseudo-code};
    \node[anchor=north west, align=left, font=\ttfamily\fontsize{7.5}{9}\selectfont]
        at (3.53,3.83) {
        \phantom{X}\\
        \# F\phantom{alpha}: hidden feature\\
        \# alpha\phantom{F}: user-controlled detail strength\\
        \# sigma\_t: current noise level\\
        \# kappa, rho: hyperparameters\\[9pt]
        lambda = alpha * scheduler(sigma\_t, kappa, rho)\\
        P = F ** 2\\
        P = (1 - lambda) * P + lambda * P.mean(dim=-1, keepdim=True)\\
        return sign(F) * sqrt(P)};
\end{tikzpicture}%
    }
    \vspace{-1mm}
    \caption{Illustration of the T$\flat$ pseudo-code corresponding to \cref{eq:analysis-density,eq:interpolate,eq:revert,eq:scheduler}.
    T$\flat$ is injected into the residual branch of each DiT block.}
    \label{fig:sdc}
    \vspace{-0.5mm}
\end{figure}

\section{Simple Detail Control in Diffusion Transformers}
\label{sec:sdc}
To mimic the above observed patterns, we propose a simple training-free mechanism named T$\flat$ that allows users to
explicitly control the detail richness of generated images through a scalar parameter $\alpha$ on DiTs.
The overall procedure is outlined in \cref{fig:sdc}.

\paragraph{Token Entropy Amplifier.}
Following the definition in \cref{eq:analysis-density}, we compute the normalized channel density $\rvp_{h,w}$ from each token feature $\mathbf{F}_{h,w}$.
The goal is to increase the entropy of $\rvp_{h,w}$.
Since entropy is strictly concave and the uniform distribution is its unique maximiser, we interpolate it towards $1/D$, which raises entropy monotonically:
\begin{equation}
    \rvp'_{i} = (1-\lambda(\alpha, \sigma_t))\,\rvp_{i} + \lambda(\alpha, \sigma_t)\,\frac{1}{D} \label{eq:interpolate},
\end{equation}
where $\lambda$ is a scalar determined by the user-specified guidance strength $\alpha$ and the noise level $\sigma_t$ of the current sampling step. 
Finally, we revert the redistributed density $\rvp'$ to amplitude space,
rescaling and restoring original per-channel signs:
\begin{equation}
    \rvq'_{i} = \operatorname{sign}(\rvq_{i})\,\|\rvq\|\sqrt{\rvp'_{i}} \label{eq:revert}.
\end{equation}

By default, the amplifier is applied to all feature tokens. 
For greater stability, we apply it only within the residual branch. 
In other words, it affects only the current block's residual update 
while leaving the accumulated hidden states from preceding blocks unchanged.

\paragraph{Time-aware Scheduler}
\begin{figure}[t]
    \centering
    \begin{minipage}[t]{0.19\textwidth}
        \centering
        \includegraphics[width=\linewidth]{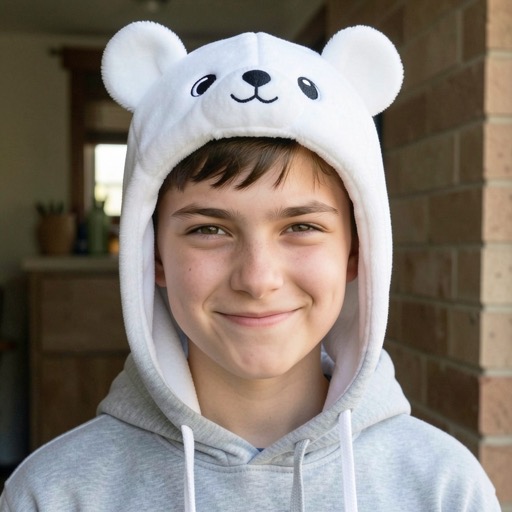}
        \par{\figurefont\scriptsize Baseline}
    \end{minipage}\hfill
    \begin{minipage}[t]{0.19\textwidth}
        \centering
        \includegraphics[width=\linewidth]{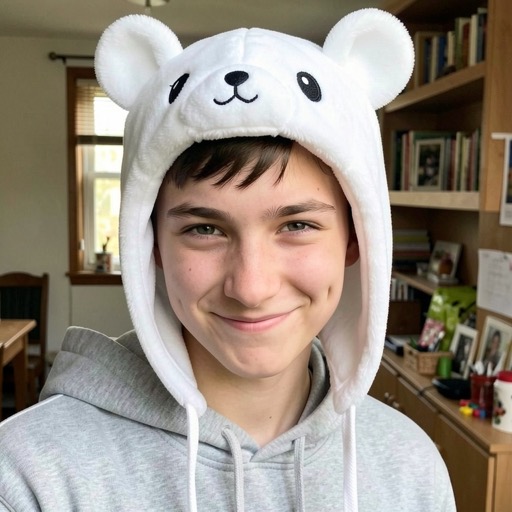}
        \par{\figurefont\scriptsize T$\flat$ on step 1}
    \end{minipage}\hfill
    \begin{minipage}[t]{0.19\textwidth}
        \centering
        \includegraphics[width=\linewidth]{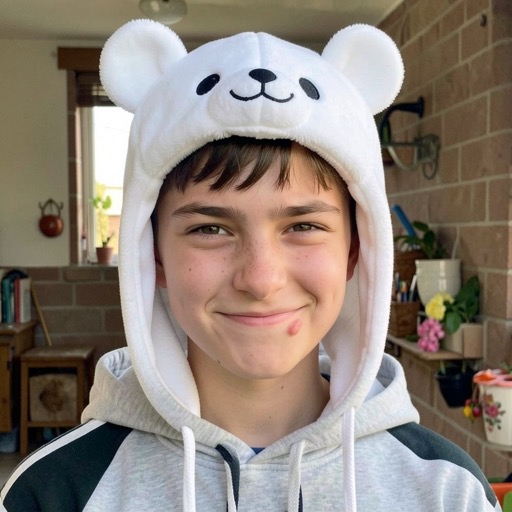}
        \par{\figurefont\scriptsize T$\flat$ on step 2}
    \end{minipage}\hfill
    \begin{minipage}[t]{0.19\textwidth}
        \centering
        \includegraphics[width=\linewidth]{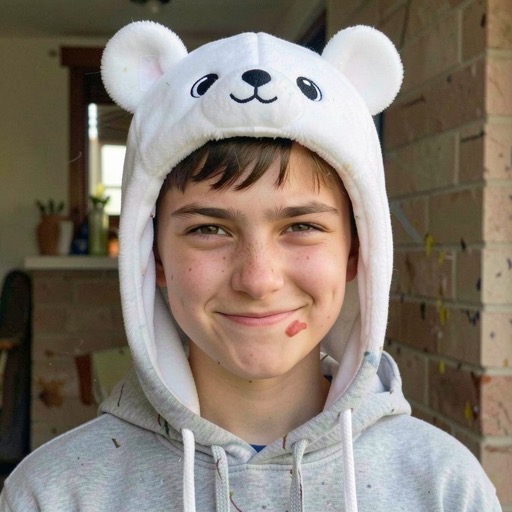}
        \par{\figurefont\scriptsize T$\flat$ on step 3}
    \end{minipage}\hfill
    \begin{minipage}[t]{0.19\textwidth}
        \centering
        \includegraphics[width=\linewidth]{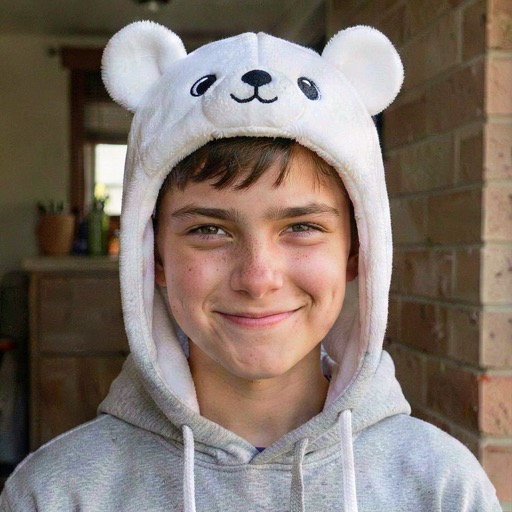}
        \par{\figurefont\scriptsize T$\flat$ on step 4 (-1)}
    \end{minipage}
    \vspace{-1mm}
    \caption{Qualitative effects of applying T$\flat$ at a single sampling step with the same positive weights using FLUX.2 Klein 9B.
    At early steps, our method tends to enhance low-frequency details, such as adding objects;
    but it may also alter the global semantic structure if applying it too early.
    At later steps, T$\flat$ introduces more localized, high-frequency details, such as subtle freckles on faces and coarse fabric textures.
    That motivates our time-aware scheduler design in \cref{sec:sdc}.}
    \label{fig:time-step-images}
    \vspace{-2mm}
\end{figure}

We model the interpolation strength \(\lambda\) in \cref{eq:interpolate} by scaling \(\alpha\) according to the current $\sigma_t$.
A flow-matching or diffusion model progressively samples from a standard Gaussian distribution ($\sigma_{t}=1$), toward a high-density data point ($\sigma_{t}=0$).
We observe that applying our operation at early timesteps alters more low-frequency components, which may undesirably affect the global semantic structure.
At late timesteps, it mainly alters high-frequency components, causing very localized changes.
A qualitative example is shown in \cref{fig:time-step-images}.
We therefore aim to concentrate the guidance on intermediate timesteps, where it primarily affects mid-frequency components that correspond to meaningful details.
Specifically, we define
\begin{equation}
    \lambda(\alpha, \sigma_t) = \alpha
\left(\frac{\sigma_t}{\rho}\right)^{\kappa\rho}
\left(\frac{1-\sigma_t}{1-\rho}\right)^{\kappa(1-\rho)}
\label{eq:scheduler},
\end{equation}
where $\rho\in(0,1)$ sets the peak location and $\kappa$ its concentration. 
Thus, $\lambda$ reaches its maximum $\alpha$ at $\sigma_t=\rho$ and remains below $\alpha$ elsewhere.

\section{Experiment}
\label{sec:experiment}
In this section, we first demonstrate the performance of T$\flat$ on three DiTs in \cref{sec:detail-control-results}.
Then, we compare T$\flat$ with other related methods in \cref{sec:comparison}.
Finally, we provide further discussion of the choice of density formulation, 
the time-wise effect, and the limitations in \cref{sec:further-discussion}.

\paragraph{Models and implementation details}
We use three representative diffusion transformers: FLUX.2 Klein 9B Base, its distilled variant FLUX.2 Klein 9B \citep{flux2klein}, and Qwen-Image (20B) \citep{qwenimage}.
The Base and Qwen models use 25 inference steps with a classifier-free guidance (CFG) scale of 4, whereas the distilled model uses 4 steps without CFG.
All models generate $768\times768$ images.
For the time-aware scheduler, we set $\kappa=40$ for all models and $\rho=0.9$, $0.7$, and $0.8$ for the distilled, Base, and Qwen models.
More details are provided in \cref{app:implementation-details}.

\paragraph{Evaluation Metrics}
To assess image quality, we use CLIP \citep{clip} (Long-CLIP-L, ViT-L/14) and HPSv3 \citep{hpsv3} (based on Qwen2-VL-7B) to evaluate text-image alignment and human preference.
We quantify image detail using four metrics:
edge strength (mean gradient magnitude), sharpness (Laplacian variance),
spectral edge frequency $f_{95}$ (radial frequency enclosing 95\% of Hann-windowed Fourier energy, in cycles per pixel),
and wavelet detail energy \citep{wavelet}.
Detailed definitions of them are provided in \cref{app:detail-metrics}.
Higher values across these metrics indicate stronger detail components.
We assess structure preservation by comparing generated images and the baseline,
using DINO \citep{dinov2} cosine distance between the CLS tokens and LPIPS \citep{lpips} computed after downsampling both images to $128\times128$.

\subsection{Detail Control Results}
\label{sec:detail-control-results}

\begin{table}[!t]
    \centering
    \resizebox{\textwidth}{!}{%
    \renewcommand{\arraystretch}{1.0}
    \begin{tabular}{l  cc  cccc  cc}
        \toprule
        \multirow{2}{*}{Input $\alpha$} & \multicolumn{2}{c}{Image Quality}
        & \multicolumn{4}{c}{Detail Statistics}
        & \multicolumn{2}{c}{Structure Difference} \\
        \cmidrule(lr){2-3}\cmidrule(lr){4-7}\cmidrule(lr){8-9}
        & CLIP$\uparrow$  & HPSv3$\uparrow$ & Edge & Sharpness & $f_{95}$ & Wavelet$$ & DINO & LPIPS$_{128}$ \\
        \midrule
        \rowcolor{lightblue}
        \multicolumn{9}{l}{\rule[-0.6ex]{0pt}{2.7ex}Qwen Image} \\
        $-0.15$ & 0.2771 & 11.212 & 0.0170 & 0.0048 & 0.0336 & 0.0196 & 0.1146 & 0.1517 \\
        $-0.1$ & \textbf{0.2774} & 11.474 & 0.0171 & 0.0051 & 0.0357 & 0.0209 & 0.0922 & 0.1251 \\
        $-0.05$ & \underline{0.2773} & 11.749 & 0.0173 & 0.0055 & 0.0381 & 0.0227 & 0.0645 & 0.0846 \\
        \rowcolor{lightgray}
        $0$ (baseline) & 0.2769 & \underline{12.011} & 0.0182 & 0.0063 & 0.0435 & 0.0268 & 0.0000 & 0.0000 \\
        $+0.05$ & 0.2765 & \textbf{12.013} & 0.0194 & 0.0072 & 0.0494 & 0.0317 & 0.0544 & 0.0713 \\
        $+0.1$ & 0.2762 & 11.880 & 0.0203 & 0.0076 & 0.0536 & 0.0358 & 0.0828 & 0.1110 \\
        $+0.15$ & 0.2759 & 11.603 & 0.0211 & 0.0079 & 0.0575 & 0.0402 & 0.1133 & 0.1524 \\
        $\nearrow_{\alpha}$ & - & - & \textcolor{green!40!black}{$\checkmark$} & \textcolor{green!40!black}{$\checkmark$} & \textcolor{green!40!black}{$\checkmark$} & \textcolor{green!40!black}{$\checkmark$} & - & - \\
        \midrule
        \rowcolor{lightgreen}
        \multicolumn{9}{l}{\rule[-0.6ex]{0pt}{2.7ex}FLUX.2 Klein 9B} \\
        $-0.03$ & \underline{0.2748} & 11.908 & 0.0313 & 0.0192 & 0.0553 & 0.0317 & 0.0532 & 0.1007 \\
        $-0.02$ & 0.2747 & 11.986 & 0.0322 & 0.0200 & 0.0566 & 0.0322 & 0.0446 & 0.0857 \\
        $-0.01$ & \textbf{0.2749} & 12.097 & 0.0335 & 0.0210 & 0.0585 & 0.0331 & 0.0315 & 0.0601 \\
        \rowcolor{lightgray}
        $0$ (baseline) & \underline{0.2748} & 12.271 & 0.0360 & 0.0230 & 0.0634 & 0.0354 & 0.0000 & 0.0000 \\
        $+0.01$ & 0.2745 & \textbf{12.434} & 0.0403 & 0.0265 & 0.0723 & 0.0391 & 0.0403 & 0.0719 \\
        $+0.02$ & 0.2738 & \underline{12.315} & 0.0448 & 0.0304 & 0.0823 & 0.0429 & 0.0759 & 0.1309 \\
        $+0.03$ & 0.2723 & 11.948 & 0.0500 & 0.0355 & 0.0952 & 0.0476 & 0.1178 & 0.1924 \\
        $\nearrow_{\alpha}$ & - & - & \textcolor{green!40!black}{$\checkmark$} & \textcolor{green!40!black}{$\checkmark$} & \textcolor{green!40!black}{$\checkmark$} & \textcolor{green!40!black}{$\checkmark$} & - & - \\
        \midrule
        \rowcolor{lightpink}
        \multicolumn{9}{l}{\rule[-0.6ex]{0pt}{2.7ex}FLUX.2 Klein Base 9B} \\
        $-0.0045$ & 0.2732 & 8.535 & 0.0275 & 0.0214 & 0.0511 & 0.0330 & 0.0743 & 0.1267 \\
        $-0.003$ & 0.2741 & 9.586 & 0.0294 & 0.0231 & 0.0560 & 0.0353 & 0.0455 & 0.0780 \\
        $-0.0015$ & 0.2745 & 10.397 & 0.0317 & 0.0247 & 0.0615 & 0.0377 & 0.0214 & 0.0340 \\
        \rowcolor{lightgray}
        $0$ (baseline) & \underline{0.2750} & 11.041 & 0.0347 & 0.0268 & 0.0682 & 0.0407 & 0.0000 & 0.0000 \\
        $+0.0015$ & \textbf{0.2751} & 11.405 & 0.0386 & 0.0299 & 0.0762 & 0.0434 & 0.0189 & 0.0284 \\
        $+0.003$ & \underline{0.2750} & \textbf{11.520} & 0.0424 & 0.0336 & 0.0845 & 0.0455 & 0.0394 & 0.0624 \\
        $+0.0045$ & 0.2749 & \underline{11.480} & 0.0465 & 0.0382 & 0.0948 & 0.0477 & 0.0616 & 0.0978 \\
        $\nearrow_{\alpha}$ & - & - & \textcolor{green!40!black}{$\checkmark$} & \textcolor{green!40!black}{$\checkmark$} & \textcolor{green!40!black}{$\checkmark$} & \textcolor{green!40!black}{$\checkmark$} & - & - \\
        \bottomrule
    \end{tabular}%
    }
    \vspace{-1mm}
    \caption{T$\flat$ results across three diffusion transformer models and three metric groups over a range of $\alpha$ values.
    $\nearrow_{\alpha}$ indicates a monotonic increase with respect to the given $\alpha$s.
    T$\flat$ provides continuous detail control while maintaining image quality and causing only small structural changes.}
    \label{tab:sdc}
    \vspace{-2mm}
\end{table}

\input{figures/fig_detail_examples.tex}

We conduct the experiment on 800 prompts obtained by combining the HPDv2 \citep{hpsv2} and HPDv3 \citep{hpsv3} test sets, 
since these prompts span diverse categories and complexity levels. 
We evaluate T$\flat$ over a reasonable range of $\alpha$ values for each model and report the results in \cref{tab:sdc}. 
Compared with the baseline ($\alpha=0$), T$\flat$ consistently increases or decreases image detail metrics as $\alpha$ becomes more positive or negative, as indicated by all four averaged detail metrics
For image quality assessment, CLIP scores barely change with $\alpha$, indicating that T$\flat$ largely preserves text-image alignment.
HPSv3 scores even improve with slightly positive $\alpha$, suggesting that a moderate increase in detail may benefit human preference. 
DINO and LPIPS$_{128}$ increase with $|\alpha|$,
but they remain relatively low within the evaluated ranges, indicating limited structural changes.
Some qualitative examples are illustrated in \cref{fig:detail-control-examples}.

\subsection{Comparison with others}
\label{sec:comparison}

\begin{table}[t]
    \centering
    \resizebox{\textwidth}{!}{%
    \renewcommand{\arraystretch}{1.05}
    \begin{tabular}{l@{\hspace{5.5pt}} c@{\hspace{10pt}} cc  cccc  cc}
        \toprule
        \multirow{2}{*}{Method} & \multirow{2}{*}{\shortstack{Latency\\ over \\ Baseline$\downarrow$}} & \multicolumn{2}{c}{Image Quality}
        & \multicolumn{4}{c}{Detail Statistics}
        & \multicolumn{2}{c}{Structure Difference} \\
        \cmidrule(lr){3-4}\cmidrule(lr){5-8}\cmidrule(lr){9-10}
        & & CLIP$\uparrow$ & HPSv3$\uparrow$ & Edge$\uparrow$ & Sharpness$\uparrow$ & $f_{95}\uparrow$ & Wavelet$\uparrow$ & DINO$\downarrow$ & LPIPS$_{128}\downarrow$ \\
        \midrule
        \rowcolor{lightgray}
        Baseline & 1.00$\times$ & \underline{0.2750} & 11.0413 & 0.0347 & 0.0268 & 0.0682 & 0.0407 & \textcolor{gray}{0.0000} & \textcolor{gray}{0.0000} \\
        SAG & 1.56$\times$ & 0.2740 & 11.0230 & \underline{0.0389} & 0.0241 & \underline{0.0776} & \textbf{0.0491} & \textcolor{gray}{0.0690} & \textcolor{gray}{0.1286} \\
        PAG & 1.53$\times$ & 0.2743 & 10.9532 & 0.0363 & \underline{0.0315} & 0.0761 & 0.0464 & \textcolor{gray}{0.0663} & \textcolor{gray}{\underline{0.1012}} \\
        SEG & 1.53$\times$ & 0.2749 & 10.5784 & 0.0372 & 0.0293 & 0.0713 & 0.0436 & \textcolor{gray}{0.1081} & \textcolor{gray}{0.1763} \\
        TPG & 1.53$\times$ & 0.2738 & 10.5830 & 0.0336 & 0.0218 & 0.0606 & 0.0385 & \textcolor{gray}{0.0752} & \textcolor{gray}{0.1303} \\
        DG & 1.53$\times$ & 0.2747 & 10.8166 & 0.0338 & 0.0252 & 0.0702 & 0.0413 & \textcolor{gray}{0.0881} & \textcolor{gray}{0.1503} \\
        SSG & 1.58$\times$ & 0.2742 & 10.2274 & 0.0288 & 0.0132 & 0.0533 & 0.0398 & \textcolor{gray}{0.0993} & \textcolor{gray}{0.1533} \\
        FreSca & \underline{1.03$\times$} & \textbf{0.2752} & \underline{11.2964} & 0.0379 & 0.0294 & 0.0763 & 0.0459 & \textcolor{gray}{0.0719} & \textcolor{gray}{0.1194} \\
        FDG & \textbf{1.02}$\bm{\times}$ & 0.2744 & 10.9110 & 0.0368 & 0.0261 & 0.0652 & 0.0414 & \textcolor{gray}{0.1774} & \textcolor{gray}{0.2724} \\
        HiGS & \textbf{1.02}$\bm{\times}$ & 0.2749 & 11.1414 & 0.0360 & 0.0277 & 0.0707 & 0.0418 & \textcolor{gray}{\underline{0.0637}} & \textcolor{gray}{0.1026} \\
        \rowcolor{lightyellow}
        \textbf{T$\bm{\flat}$ (Ours)} & \underline{1.03$\times$} & 0.2749 & \textbf{11.4796} & \textbf{0.0465} & \textbf{0.0382} & \textbf{0.0948} & \underline{0.0477} & \textcolor{gray}{\textbf{0.0616}} & \textcolor{gray}{\textbf{0.0978}} \\
        \bottomrule
    \end{tabular}%
    }
    \vspace{-1mm}
    \caption{Comparison with training-free detail enhancement methods on FLUX.2 Klein 9B Base. T$\flat$ achieves the strongest detail statistics and quality with smallest structural changes.
    It adds negligible inference time compared to the baseline.}
    \label{tab:comparison}
    \vspace{-2mm}
\end{table}

\begin{figure}[t]
    \centering
    \begin{tikzpicture}
        \node[inner sep=0] (comparison-row)
            {\includegraphics[width=\textwidth]{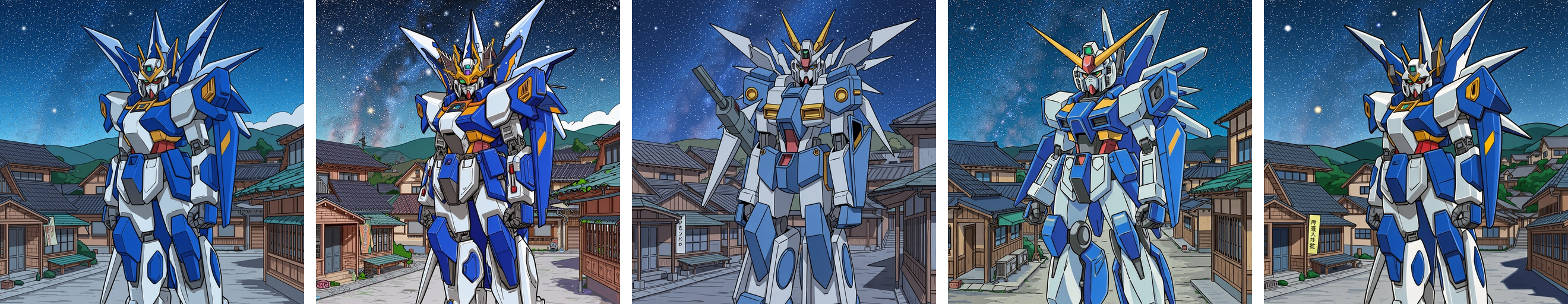}};
        \foreach \position/\method/\boxcolor in {
            0.09697/Baseline/lightgray,
            0.29848/\textbf{T$\bm{\flat}$ (Ours)}/lightyellow,
            0.50000/DG/lightblue,
            0.70152/FreSca/modelaqua,
            0.90303/SAG/modellavender}
            \node[anchor=south, yshift=2pt, inner sep=2.5pt, minimum width=0.19\textwidth, minimum height=12.5pt,
                rounded corners=1.5pt, fill=\boxcolor,
                font={\figurefont\fontsize{7.5}{9}\selectfont}]
                at ($(comparison-row.north west)!\position!(comparison-row.north east)$) {\method};
    \end{tikzpicture}
    \par\vspace{-0.5mm}
    {\color{black!70}\figurefont\fontsize{7.5}{9}\selectfont\itshape
    A portrait of an anime mecha robot with a Japanese town background and a starred night sky.}
    \par\smallskip
    \includegraphics[width=\textwidth]{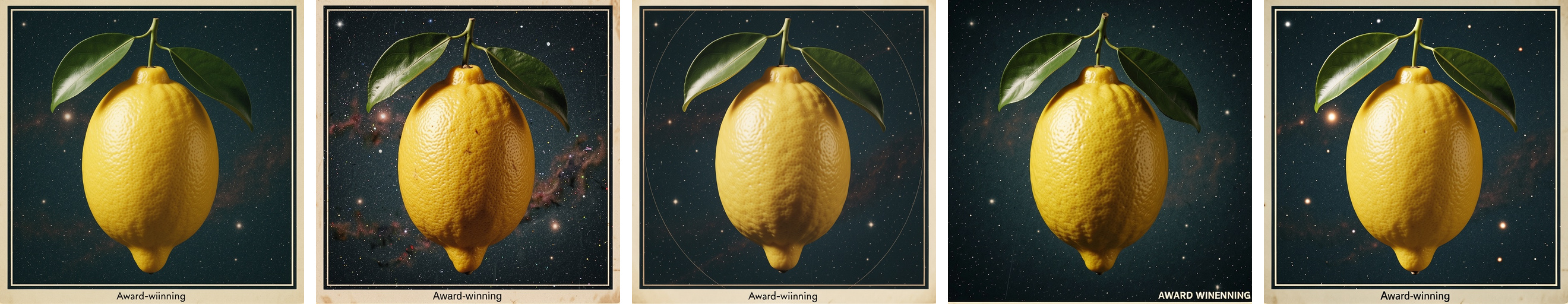}
    \par\vspace{-0.5mm}
    {\color{black!70}\figurefont\fontsize{7.5}{9}\selectfont\itshape
    An award-winning portrait of a lemon in a muted, space age style reminiscent of the 1930s.}
    \par\smallskip
    \includegraphics[width=\textwidth]{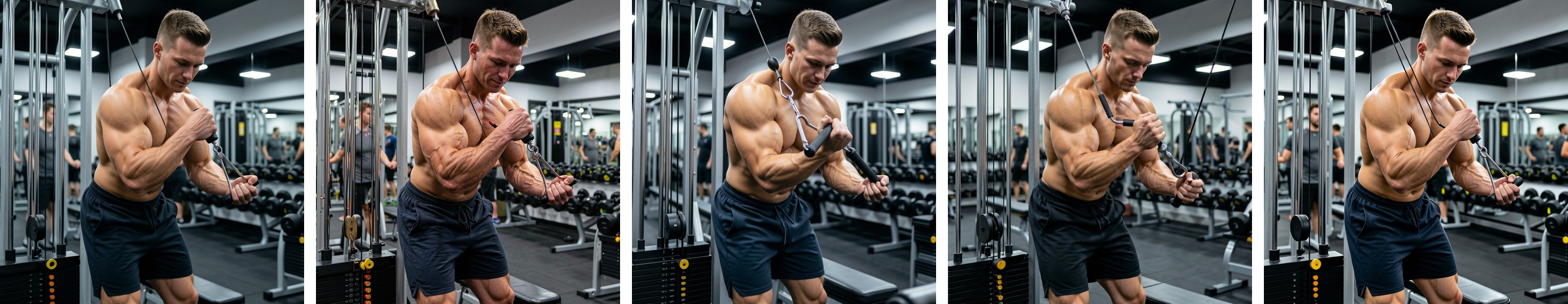}
    \par\vspace{-0.5mm}
    {\color{black!70}\figurefont\fontsize{7.5}{9}\selectfont\itshape
    Muscular man working out biceps with cable machine in gym.}
    \vspace{-1mm}
    \caption{Comparison with the best-performing competing methods in \cref{tab:comparison}.
    T$\flat$ better enhances local details over the baseline.
    Zoom in to better view the differences. More examples in \cref{app:more-comparisons}.}
    \label{fig:comparison}
    \vspace{-3mm}
\end{figure}

Several existing training-free methods may also enhance image details. 
First, we compare against methods that introduce additional guidance, 
including SAG \citep{sag}, PAG \citep{pag}, SEG \citep{seg}, TPG \citep{tpg}, DG \citep{dg}, and SSG \citep{ssg}. 
These methods employ CFG-like operations to steer the sampling process away from negative predictions constructed 
by perturbing intermediate features in different ways. 
We also compare with FreSca \citep{fresca} and FDG \citep{fdg}, which apply different guidance weights to the high and low frequency components of the CFG term. 
Finally, we compare with HiGS \citep{higs}, 
which enhances details using a weighted average of historical predictions.

For a fair comparison, we apply all methods to FLUX.2 Klein 9B Base 
using identical settings and the same prompt set. 
We progressively increase the guidance strength of each method until its structural deviation from the baseline exceeds that of ours. 
Even at this point, our T$\flat$ achieves higher detail metrics. 
This indicates that our method provides more effective, detail-focused control.
In terms of image quality, while all the methods maintain a similar level of CLIP score,
T$\flat$ delivers the highest HPSv3 preference.
In addition, most guidance methods require about $0.5\times$ more inference computation,
while T$\flat$ only adds a negligible overhead of $3\%$.
We report the results and the added computation over baseline in \cref{tab:comparison}.
A qualitative comparison is shown in \cref{fig:comparison}.

\subsection{Further discussion}
\label{sec:further-discussion}

\newcommand{\timestepenergyplot}[5]{%
    \begin{tikzpicture}
        \begin{axis}[
            width=1.15\linewidth,
            height=0.9\linewidth,
            font=\figurefont,
            title={#1},
            title style={font=\figurefont\scriptsize, yshift=-8pt},
            ybar,
            bar width=5pt,
            ymin=#2,
            ymax=#3,
            ylabel={#4},
            ylabel style={font=\figurefont\scriptsize},
            tick label style={font=\figurefont\scriptsize},
            symbolic x coords={low,mid-low,middle,mid-high,high},
            xtick=data,
            x tick label style={rotate=45, anchor=east, font=\figurefont\scriptsize},
            axis x line*=bottom,
            axis y line*=left,
            axis line style={black!70},
            tick align=outside,
            enlarge x limits=0.12,
            grid=major,
            grid style={draw=gray!18},
        ]
            \addplot[fill=blue!55, draw=blue!70!black] coordinates {#5};
            \draw[black!70, thin] (axis cs:low,0) -- (axis cs:high,0);
        \end{axis}
    \end{tikzpicture}%
}

\begin{figure}[t]
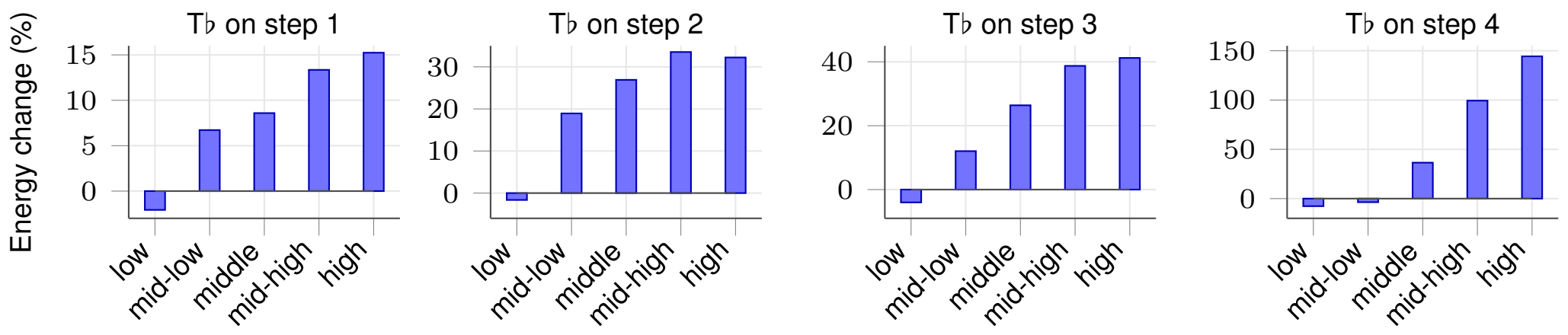

    \centering
    \begin{subfigure}[t]{0.245\textwidth}
        \centering
        \timestepenergyplot{T$\flat$ on step 1}{-3}{16}{Energy change (\%)}{
            (low,-2.07) (mid-low,6.72) (middle,8.59)
            (mid-high,13.35) (high,15.24)
        }
    \end{subfigure}%
    \begin{subfigure}[t]{0.245\textwidth}
        \centering
        \timestepenergyplot{T$\flat$ on step 2}{-6}{35}{}{
            (low,-1.63) (mid-low,18.93) (middle,26.91)
            (mid-high,33.53) (high,32.23)
        }
    \end{subfigure}%
    \begin{subfigure}[t]{0.245\textwidth}
        \centering
        \timestepenergyplot{T$\flat$ on step 3}{-9}{45}{}{
            (low,-4.01) (mid-low,12.04) (middle,26.38)
            (mid-high,38.72) (high,41.21)
        }
    \end{subfigure}%
    \begin{subfigure}[t]{0.245\textwidth}
        \centering
        \timestepenergyplot{T$\flat$ on step 4}{-20}{155}{}{
            (low,-7.65) (mid-low,-3.60) (middle,36.46)
            (mid-high,99.38) (high,144.33)
        }
    \end{subfigure}
    \vspace{-4mm}
    \caption{Percentage change in spectral energy of different frequency bands 
    when T$\flat$ is applied at a single sampling step using the same weight with FLUX.2 Klein 9B. }
    \label{fig:time-step-energy}
    \vspace{-2mm}
\end{figure}

\let\timestepenergyplot\relax

\begin{figure}[t]
    \centering
    \makebox[\textwidth][c]{%
    \begin{minipage}[t]{0.35\textwidth}
    \centering
    \begin{minipage}[t][0.29\textwidth][t]{\linewidth}
    \centering
    \begin{tikzpicture}
        \begin{axis}[
            width=0.97\linewidth,
            height=0.738\linewidth,
            font=\figurefont,
            ybar,
            bar width=10pt,
            ymin=0,
            ymax=0.18,
            ytick={0,0.05,0.10,0.15},
            scaled y ticks=false,
            y tick label style={
                /pgf/number format/fixed,
                /pgf/number format/precision=2,
            },
            xtick={1,2,3,4},
            xmin=0.5,
            xmax=4.5,
            xlabel={T$\flat$ applying step},
            ylabel={LPIPS$_{128}$},
            label style={font=\figurefont\scriptsize},
            tick label style={font=\figurefont\scriptsize},
            xticklabel style={yshift=3pt},
            xlabel style={yshift=3pt},
            axis x line*=bottom,
            axis y line*=left,
            axis line style={black!70},
            tick align=outside,
            grid=major,
            grid style={draw=gray!18},
        ]
            \addplot[fill=orange!70, draw=orange!85!black] coordinates {
                (1,0.1634)
                (2,0.1657)
                (3,0.0916)
                (4,0.0546)
            };
        \end{axis}
    \end{tikzpicture}
    \end{minipage}
    \vspace{-6mm}
    \captionof{figure}{Structural difference from the baseline when T$\flat$ is applied at a
    single sampling step.}
    \label{fig:time-step-lpips}
    \end{minipage}\hspace{0.02\textwidth}%
    \begin{minipage}[t]{0.62\textwidth}
        \centering
        \begin{minipage}[t][0.31\textwidth][t]{\linewidth}
            \centering
            \pgfplotsset{scheduleraxisstyle/.style={
    width=1.15\linewidth,
    height=0.81\linewidth,
    font=\figurefont,
    xmin=0, xmax=1,
    x coord trafo/.code={\pgfmathparse{##1^1.5}},
    x coord inv trafo/.code={\pgfmathparse{##1^0.6666667}},
    ymin=0, ymax=0.032,
    xtick={0,0.5,0.7,0.8,0.9,1.0},
    ytick={0.03},
    yticklabels={$\alpha$},
    scaled y ticks=false,
    xlabel={$\sigma_t$},
    ylabel={Guidance ($\lambda$)},
    label style={font=\figurefont\scriptsize},
    ylabel style={font=\figurefont\scriptsize, xshift=0pt, yshift=-8pt},
    tick label style={font=\figurefont\scriptsize},
    xticklabel style={yshift=3pt},
    xlabel style={yshift=3pt},
    legend pos=north west,
    legend style={font=\figurefont\tiny, draw=none, inner sep=1pt, row sep=-1pt,
        xshift=-2pt, yshift=-2pt},
    legend image post style={xscale=0.5},
    axis x line*=bottom,
    axis y line*=left,
    axis line style={black!70},
    tick align=outside,
    grid=major,
    grid style={draw=gray!18},
}}

\begin{minipage}[t]{0.50\linewidth}
    \centering
    \begin{tikzpicture}
        \begin{axis}[scheduleraxisstyle]
            \addplot[blue!70!black, very thick, domain=0:1, samples=201]
                {0.03*(x/0.8)^32*((1-x)/0.2)^8};
            \addlegendentry{$\rho=0.8$}
            \addplot[orange!85!black, very thick, dashed, domain=0:1, samples=201]
                {0.03*(x/0.7)^28*((1-x)/0.3)^12};
            \addlegendentry{$\rho=0.7$}
            \addplot[green!55!black, very thick, dotted, domain=0:1, samples=201]
                {0.03*(x/0.9)^36*((1-x)/0.1)^4};
            \addlegendentry{$\rho=0.9$}
            \draw[dashed, gray!70] (axis cs:0,0.03) -- (axis cs:1,0.03);
        \end{axis}
    \end{tikzpicture}
\end{minipage}%
\begin{minipage}[t]{0.50\linewidth}
    \centering
    \begin{tikzpicture}
        \begin{axis}[scheduleraxisstyle]
            \addplot[blue!70!black, very thick, domain=0:1, samples=201]
                {0.03*(x/0.8)^32*((1-x)/0.2)^8};
            \addlegendentry{$\kappa=40$}
            \addplot[orange!85!black, very thick, dashed, domain=0:1, samples=201]
                {0.03*(x/0.8)^16*((1-x)/0.2)^4};
            \addlegendentry{$\kappa=20$}
            \addplot[green!55!black, very thick, dotted, domain=0:1, samples=201]
                {0.03*(x/0.8)^64*((1-x)/0.2)^16};
            \addlegendentry{$\kappa=80$}
            \draw[dashed, gray!70] (axis cs:0,0.03) -- (axis cs:1,0.03);
        \end{axis}
    \end{tikzpicture}
\end{minipage}
        \end{minipage}
        \vspace{-6mm}
        \captionof{figure}{Time-aware scheduler curves. 
        Left: varying $\rho$ shifts the peak of the curve. 
        Right: varying $\kappa$ controls the width of the curve.
        The peak is always equal to $\alpha$. 
        }
        \label{fig:scheduler-curve}
    \end{minipage}
    }
    \vspace{-4mm}
\end{figure}

\paragraph{Time-aware effects of T$\bm{\flat}$.}
We apply T$\flat$ at one sampling step at a time with a fixed guidance weight $\alpha>0$ using the four-step FLUX.2 Klein 9B model.
In \cref{fig:time-step-energy}, we report the percentage change in spectral energy across five frequency bands, whose definitions are provided in \cref{app:time-aware-effects}.
Early interventions distribute the energy change relatively evenly across frequencies, 
whereas later interventions increasingly amplify high-frequency energy. 
\Cref{fig:time-step-lpips} shows that early interventions cause larger structural deviations from the baseline. 
A qualitative example is provided in \cref{fig:time-step-images}.
Together, these observations motivate the time-aware scheduler in \cref{sec:sdc}, 
which emphasizes guidance at intermediate sampling steps. 
\Cref{fig:scheduler-curve} visualizes the scheduler curves: 
$\rho$ determines the peak location, while $\kappa$ controls the width. 
A larger $\kappa$ produces a sharper and more concentrated schedule. 
Both parameters can be adjusted to accommodate different user preferences.

\paragraph{Why not define density using absolute magnitudes?}
In \cref{eq:analysis-density}, we define channel density using squared magnitudes.
A straightforward alternative is to use absolute magnitudes. 
We repeat the analysis in \cref{fig:observation} with this definition and report the results in \cref{tab:magnitude-comparison}. 
Compared with squared magnitudes, absolute ones yield lower entropy-increase ratios. 
We attribute this difference to the amplification effect of squaring. 
Squaring emphasizes changes among higher-magnitude channels while suppressing contributions from lower ones. 
This reveals that detail instructions redistribute density primarily among highly active channels.
This observation is consistent with prior findings on massive activations,
where tokens are often sparse and dominated by few channels \citep{dg,ditf,calibatt}.
The effect is more pronounced for models with sparser tokens.

\paragraph{T$\bm{\flat}$ relies on the model capabilities.}
We further evaluate T$\flat$ on the other DiT families.
As reported in \cref{tab:appendix-sdc-families}, T$\flat$ retains monotonic control over most detail metrics but is less effective than on stronger FLUX and Qwen models,
and image quality and structural similarity decline faster for the same change.
We hypothesize that stronger models learn a more densely populated feature space and tolerate token perturbations better, 
whereas perturbing weaker models may more easily push features outside the learned distribution, degrading image quality.
This dependence on the inherent capability of the pretrained model represents a limitation of T$\flat$ as a training-free method.

\paragraph{Failure cases and future work.}
T$\flat$ operates within a limited effective range and may come with some side effects.
Additional examples are provided in \cref{app:qualitative}.
A fixed pushing toward a uniform target is unlikely to be the optimal direction for every token, 
but characterizing a per-token optimal direction is hard for a training-free method.
We hope this observation motivates future work that enables fine-grained control over detail, which is further discussed in \cref{app:future-work}.

\section{Conclusion}
We analyze how detail-guided prompts alter feature tokens across diffusion transformers and identify a consistent pattern.
Motivated by this observation, we propose T$\flat$, a simple training-free method for controlling detail richness with a user-specified scalar.
Our work provides insight into the mechanisms behind prompt engineering and an effective method for precise detail control.

\bibliography{iclr2027_conference}
\bibliographystyle{iclr2027_conference}

\clearpage
\appendix
\section*{Appendix}

\section{More about the observation experiment}
\label{app:more_obv}

\subsection{Detail instruction search}
\label{app:phrase-search}
Before the analysis in \cref{sec:analysis}, we search for a model-specific prompt suffix that consistently elicits more visual details from each model.
Specifically, we first did a detail-guided prompt search for each model to get phrases that can enable more detail, measured by the Fourier high-frequency energy ratio.
We build a phrase bank from an LLM: e.g. “ultra detailed”, “fine textures”, “micro-contrast”, and selected phrases as a suffix for each prompt and did a voting based on the increase of the detail measurement.
Each model gets the highest-voted phrases to use as a suffix.
In the end, we get the phrases for each model and use them as suffix.
Then we input the same image VAE latent to the diffusion transformer with different text conditions, one with the original prompt and the other with the detail-guided suffixes.

After the search, we use the following model-specific suffixes in the final analysis experiments:
``hyper realistic, ultra detailed, high resolution'' for FLUX.2,
``intricate surface detail, fine textures, ultra detailed'' for SD3.5,
and ``micro-contrast, studio lighting, fine textures'' for Z-Image.

During this prompt search, we also have some interesting findings. 
Effective detail-guided suffixes are strongly model-dependent.
For example, ``ultra detailed'' increases the FFT-based detail score for FLUX.2 but decreases it for Z-Image.
Across our search, the preferred suffixes vary more across models than across input prompts, 
indicating that detail instructions are more model-dependent than prompt-dependent.
Longer suffixes provide no consistent benefit and can even reduce detail; empirically, two to four phrases are sufficient.
It suggests that prompt engineering is not a very reliable way to control details, and T$\flat$ provides a more robust alternative.

\subsection{Comparison of entropy trends between absolute and squared magnitudes}
\label{app:entropy-magnitude-trends}
We further examine how our observation changes when token density is defined using absolute rather than squared channel magnitudes.
The complete comparison is reported in \cref{tab:magnitude-comparison}.
The absolute-magnitude formulation yields lower entropy-increase agreement ratios, especially for the sparser representations of SD3.5 and Z-Image.
As discussed in \cref{sec:further-discussion}, squaring emphasizes changes among highly active channels and suppresses weaker responses, making the density redistribution under detail instructions more evident.

\begin{table}[!b]
    \centering
    \resizebox{0.8\textwidth}{!}{%
        \renewcommand{\arraystretch}{1.0}
        \begin{tabular}{llcccc}
            \toprule
            Model & Magnitude & $\overline{H}(\rvp)$ & $\overline{H}(\hat{\rvp})$ & $A_{\mathrm{all}}$ & $A_{\mathrm{prompt}}$ \\
            \midrule
            \multirow{2}{*}{FLUX.2 Klein 9B Base}
            & \cellcolor{lightblue}square & \cellcolor{lightblue}4.463 & \cellcolor{lightblue}4.530 & \cellcolor{lightblue}78.0\% & \cellcolor{lightblue}100.0\% \\
            & absolute & 7.824 & 7.827 & 68.2\% & 100.0\% \\
            \midrule
            \multirow{2}{*}{FLUX.2 Klein 9B}
            & \cellcolor{lightblue}square & \cellcolor{lightblue}4.307 & \cellcolor{lightblue}4.433 & \cellcolor{lightblue}83.7\% & \cellcolor{lightblue}100.0\% \\
            & absolute & 7.825 & 7.830 & 68.9\% & 100.0\% \\
            \midrule
            \multirow{2}{*}{SD3.5 Large}
            & \cellcolor{lightblue}square & \cellcolor{lightblue}3.065 & \cellcolor{lightblue}3.080 & \cellcolor{lightblue}71.4\% & \cellcolor{lightblue}98.2\% \\
            & absolute & 7.150 & 7.149 & 47.6\% & 37.4\% \\
            \midrule
            \multirow{2}{*}{SD3.5 Large Turbo}
            & \cellcolor{lightblue}square & \cellcolor{lightblue}3.006 & \cellcolor{lightblue}3.027 & \cellcolor{lightblue}70.0\% & \cellcolor{lightblue}96.2\% \\
            & absolute & 7.151 & 7.152 & 59.1\% & 83.4\% \\
            \midrule
            \multirow{2}{*}{Z-Image Base}
            & \cellcolor{lightblue}square & \cellcolor{lightblue}0.057 & \cellcolor{lightblue}0.057 & \cellcolor{lightblue}46.6\% & \cellcolor{lightblue}70.8\% \\
            & absolute & 4.252 & 4.251 & 45.8\% & 36.0\% \\
            \midrule
            \multirow{2}{*}{Z-Image Turbo}
            & \cellcolor{lightblue}square & \cellcolor{lightblue}1.087 & \cellcolor{lightblue}1.096 & \cellcolor{lightblue}62.2\% & \cellcolor{lightblue}94.4\% \\
            & absolute & 6.700 & 6.700 & 43.6\% & 51.6\% \\
            \bottomrule
        \end{tabular}%
    }
    \caption{Comparison of token densities constructed from squared and absolute channel magnitudes.
    For each model, $\overline{H}(\rvp)$ and $\overline{H}(\hat{\rvp})$ denote the mean entropy without and with detail instructions, respectively.
    We also report the token-level and prompt-level entropy-increase agreement ratios, $A_{\mathrm{all}}$ and $A_{\mathrm{prompt}}$.}
    \label{tab:magnitude-comparison}
    \vspace{-2mm}
\end{table}

\subsection{Block-wise Trend}
\label{app:block-wise-trend}

We further examine how the entropy-increase agreement ratio varies across DiT blocks.
For each block, we compute the fraction of spatially corresponding token pairs whose squared-magnitude entropy increases under detail instructions, averaged over prompts, timesteps, and image tokens.
As shown in \cref{fig:block-wise-agreement}, the block-wise trends vary across models and reveal no consistent advantage for weighting specific blocks more strongly.
For both FLUX.2 variants, the last 24 single-stream blocks show a markedly stronger entropy-increase trend than the first eight double-stream blocks.
We therefore use a uniform weight across block indices in T$\flat$ as a simple default.

\newcommand{\blocktrendplot}[6]{%
    \begin{tikzpicture}
        \begin{axis}[
            width=0.98\linewidth,
            height=0.72\linewidth,
            font=\figurefont,
            title={#1},
            title style={font=\figurefont\scriptsize, yshift=-2pt, fill=#4,
                rounded corners=1.5pt, inner xsep=3pt, inner ysep=1.5pt},
            xmin=0,
            xmax=#2,
            enlarge x limits=0.03,
            ymin=0,
            ymax=100,
            xtick={0,10,20,30},
            ytick={0,25,50,75,100},
            yticklabel={\pgfmathprintnumber{\tick}\%},
            xlabel={Block index},
            ylabel={#3},
            label style={font=\figurefont\scriptsize},
            tick label style={font=\figurefont\scriptsize},
            axis x line*=bottom,
            axis y line*=left,
            axis line style={black!70},
            tick align=outside,
            grid=major,
            grid style={draw=gray!18},
            clip=false,
        ]
            \draw[black!75, dashed, line width=0.6pt]
                (axis cs:0,50) -- (axis cs:#2,50);
            \addplot[#5, line width=1pt] coordinates {#6};
        \end{axis}
    \end{tikzpicture}%
}

\begin{figure}[t]
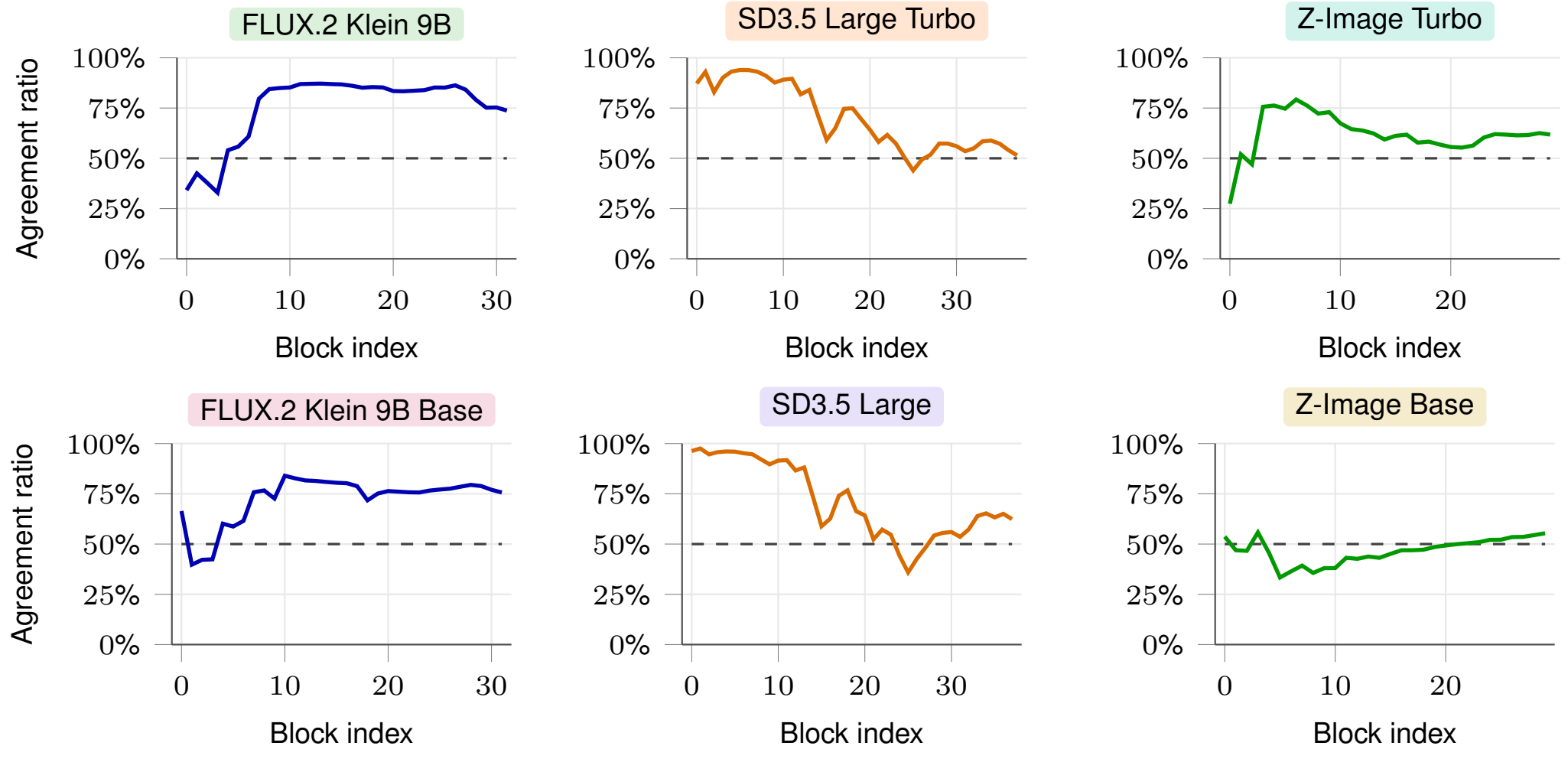

    \centering
    \begin{minipage}[t]{0.33\textwidth}
        \centering
        \blocktrendplot{FLUX.2 Klein 9B}{31}{Agreement ratio}{lightgreen}{blue!70!black}{
            (0,34.146) (1,42.497) (2,37.786) (3,32.918) (4,54.014) (5,55.791)
            (6,60.821) (7,79.636) (8,84.422) (9,84.946) (10,85.226) (11,86.926)
            (12,87.062) (13,87.153) (14,86.928) (15,86.759) (16,86.102) (17,85.118)
            (18,85.426) (19,85.252) (20,83.456) (21,83.340) (22,83.613) (23,83.905)
            (24,85.225) (25,85.172) (26,86.305) (27,84.161) (28,79.092) (29,75.187)
            (30,75.282) (31,73.753)
        }
    \end{minipage}%
    \begin{minipage}[t]{0.33\textwidth}
        \centering
        \blocktrendplot{SD3.5 Large Turbo}{37}{}{modelpeach}{orange!85!black}{
            (0,87.215) (1,92.988) (2,82.922) (3,90.087) (4,93.133) (5,93.929)
            (6,93.879) (7,93.089) (8,91.012) (9,87.672) (10,89.102) (11,89.575)
            (12,81.832) (13,83.976) (14,71.322) (15,59.115) (16,64.969) (17,74.603)
            (18,74.899) (19,69.450) (20,64.227) (21,58.154) (22,61.629) (23,57.331)
            (24,50.369) (25,43.939) (26,49.371) (27,51.775) (28,57.329) (29,57.301)
            (30,56.032) (31,53.520) (32,54.962) (33,58.433) (34,58.827) (35,57.233)
            (36,54.067) (37,51.461)
        }
    \end{minipage}%
    \begin{minipage}[t]{0.33\textwidth}
        \centering
        \blocktrendplot{Z-Image Turbo}{29}{}{modelaqua}{green!60!black}{
            (0,27.379) (1,51.971) (2,47.008) (3,75.627) (4,76.258) (5,74.689)
            (6,79.246) (7,76.185) (8,72.234) (9,72.963) (10,67.405) (11,64.514)
            (12,63.841) (13,62.366) (14,59.331) (15,61.181) (16,61.758) (17,57.787)
            (18,58.298) (19,56.852) (20,55.615) (21,55.318) (22,56.298) (23,60.338)
            (24,62.022) (25,61.774) (26,61.409) (27,61.566) (28,62.529) (29,61.811)
        }
    \end{minipage}

    \vspace{1mm}
    \begin{minipage}[t]{0.33\textwidth}
        \centering
        \blocktrendplot{FLUX.2 Klein 9B Base}{31}{Agreement ratio}{lightpink}{blue!70!black}{
            (0,66.416) (1,39.757) (2,42.168) (3,42.399) (4,60.142) (5,58.720)
            (6,61.533) (7,75.790) (8,76.676) (9,72.637) (10,83.979) (11,82.683)
            (12,81.649) (13,81.389) (14,80.910) (15,80.500) (16,80.261) (17,78.781)
            (18,71.754) (19,75.131) (20,76.362) (21,76.054) (22,75.763) (23,75.665)
            (24,76.598) (25,77.126) (26,77.557) (27,78.527) (28,79.443) (29,78.859)
            (30,76.994) (31,75.656)
        }
    \end{minipage}%
    \begin{minipage}[t]{0.33\textwidth}
        \centering
        \blocktrendplot{SD3.5 Large}{37}{}{modellavender}{orange!85!black}{
            (0,96.285) (1,97.541) (2,94.619) (3,95.719) (4,96.073) (5,95.981)
            (6,95.172) (7,94.690) (8,92.155) (9,89.663) (10,91.524) (11,91.724)
            (12,86.609) (13,88.098) (14,73.613) (15,58.850) (16,62.654) (17,74.025)
            (18,76.702) (19,66.319) (20,64.274) (21,52.335) (22,57.195) (23,54.692)
            (24,44.253) (25,35.894) (26,42.630) (27,48.171) (28,54.264) (29,55.553)
            (30,55.993) (31,53.571) (32,57.309) (33,63.921) (34,65.276) (35,63.253)
            (36,65.048) (37,62.387)
        }
    \end{minipage}%
    \begin{minipage}[t]{0.33\textwidth}
        \centering
        \blocktrendplot{Z-Image Base}{29}{}{modelsand}{green!60!black}{
            (0,53.690) (1,46.917) (2,46.637) (3,55.800) (4,45.647) (5,33.367)
            (6,36.409) (7,39.275) (8,35.637) (9,38.012) (10,38.060) (11,43.176)
            (12,42.655) (13,43.796) (14,43.160) (15,45.107) (16,46.903) (17,46.931)
            (18,47.204) (19,48.536) (20,49.290) (21,49.952) (22,50.404) (23,50.928)
            (24,52.081) (25,52.156) (26,53.464) (27,53.552) (28,54.447) (29,55.373)
        }
    \end{minipage}
    \vspace{-2mm}
    \caption{Block-wise entropy-increase agreement ratios computed from squared channel magnitudes.
    Each point is averaged over timesteps and image tokens.}
    \label{fig:block-wise-agreement}
\end{figure}

\let\blocktrendplot\relax

\subsection{Time-wise Trend}
\label{app:time-wise-trend}

We also measure the entropy-increase agreement ratio at each sampling step, averaging over prompts, blocks, and image tokens.
As shown in \cref{fig:time-wise-agreement}, the trend is relatively smooth within each model, although both FLUX.2 variants show a clearer decrease over time.
However, applying T$\flat$ at different sampling steps produces substantially different frequency and structural effects.
We therefore determine the time-aware guidance weights using the analyses in \cref{sec:further-discussion}.

\newcommand{\timetrendplot}[7]{%
    \begin{tikzpicture}
        \begin{axis}[
            width=0.96\linewidth,
            height=0.72\linewidth,
            font=\figurefont,
            title={#1},
            title style={font=\figurefont\scriptsize, yshift=-2pt, fill=#5,
                rounded corners=1.5pt, inner xsep=3pt, inner ysep=1.5pt},
            xmin=0,
            xmax=#2,
            enlarge x limits=0.03,
            ymin=0,
            ymax=100,
            xtick={#3},
            xticklabel={\pgfmathparse{int(\tick+1)}\pgfmathprintnumber{\pgfmathresult}},
            ytick={0,25,50,75,100},
            yticklabel={\pgfmathprintnumber{\tick}\%},
            xlabel={Sampling step},
            ylabel={#4},
            label style={font=\figurefont\scriptsize},
            tick label style={font=\figurefont\scriptsize},
            axis x line*=bottom,
            axis y line*=left,
            axis line style={black!70},
            tick align=outside,
            grid=major,
            grid style={draw=gray!18},
            clip=false,
        ]
            \draw[black!75, dashed, line width=0.6pt]
                (axis cs:0,50) -- (axis cs:#2,50);
            \addplot[#6, line width=1pt] coordinates {#7};
        \end{axis}
    \end{tikzpicture}%
}

\begin{figure}[t]
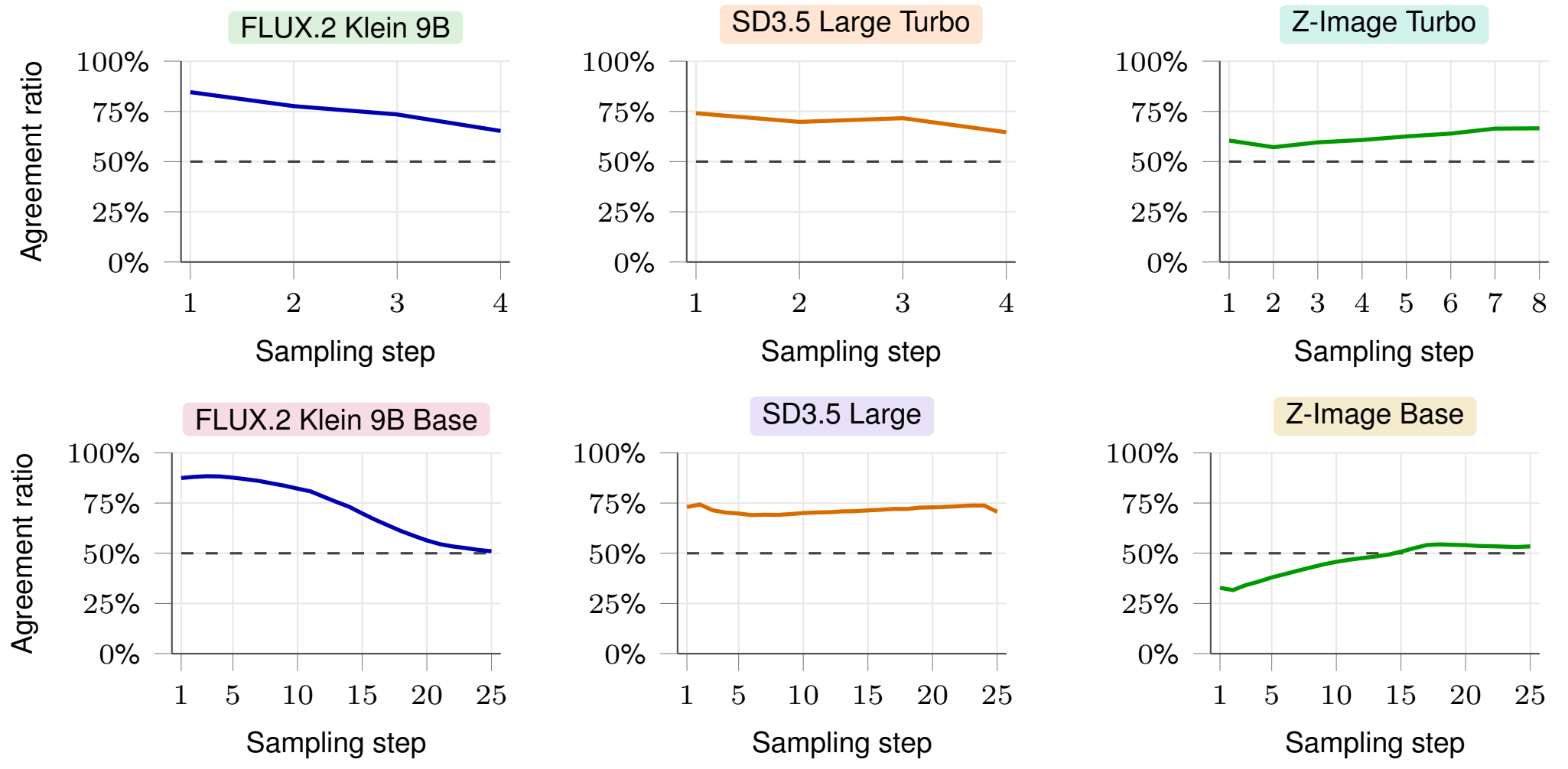

    \centering
    \begin{minipage}[t]{0.33\textwidth}
        \centering
        \timetrendplot{FLUX.2 Klein 9B}{3}{0,1,2,3}{Agreement ratio}
            {lightgreen}{blue!70!black}{
            (0,84.578) (1,77.621) (2,73.469) (3,65.259)
        }
    \end{minipage}%
    \begin{minipage}[t]{0.33\textwidth}
        \centering
        \timetrendplot{SD3.5 Large Turbo}{3}{0,1,2,3}{}
            {modelpeach}{orange!85!black}{
            (0,74.067) (1,69.772) (2,71.618) (3,64.620)
        }
    \end{minipage}%
    \begin{minipage}[t]{0.33\textwidth}
        \centering
        \timetrendplot{Z-Image Turbo}{7}{0,1,2,3,4,5,6,7}{}
            {modelaqua}{green!60!black}{
            (0,60.488) (1,57.180) (2,59.575) (3,60.762)
            (4,62.523) (5,63.963) (6,66.427) (7,66.569)
        }
    \end{minipage}

    \vspace{1mm}
    \begin{minipage}[t]{0.33\textwidth}
        \centering
        \timetrendplot{FLUX.2 Klein 9B Base}{24}{0,4,9,14,19,24}{Agreement ratio}
            {lightpink}{blue!70!black}{
            (0,87.479) (1,88.030) (2,88.372) (3,88.246) (4,87.660)
            (5,86.879) (6,86.035) (7,84.807) (8,83.626) (9,82.183)
            (10,80.840) (11,78.198) (12,75.582) (13,73.122) (14,69.843)
            (15,66.703) (16,63.884) (17,61.083) (18,58.676) (19,56.394)
            (20,54.560) (21,53.407) (22,52.588) (23,51.657) (24,50.991)
        }
    \end{minipage}%
    \begin{minipage}[t]{0.33\textwidth}
        \centering
        \timetrendplot{SD3.5 Large}{24}{0,4,9,14,19,24}{}
            {modellavender}{orange!85!black}{
            (0,73.014) (1,74.305) (2,71.401) (3,70.250) (4,69.798)
            (5,69.000) (6,69.205) (7,69.103) (8,69.498) (9,70.021)
            (10,70.306) (11,70.445) (12,70.849) (13,70.949) (14,71.340)
            (15,71.668) (16,72.083) (17,72.036) (18,72.736) (19,72.858)
            (20,73.075) (21,73.426) (22,73.783) (23,73.767) (24,70.632)
        }
    \end{minipage}%
    \begin{minipage}[t]{0.33\textwidth}
        \centering
        \timetrendplot{Z-Image Base}{24}{0,4,9,14,19,24}{}
            {modelsand}{green!60!black}{
            (0,32.795) (1,31.665) (2,34.152) (3,35.893) (4,37.966)
            (5,39.581) (6,41.290) (7,42.880) (8,44.445) (9,45.749)
            (10,46.765) (11,47.590) (12,48.379) (13,49.279) (14,50.786)
            (15,52.516) (16,54.106) (17,54.417) (18,54.201) (19,54.038)
            (20,53.619) (21,53.503) (22,53.294) (23,53.132) (24,53.433)
        }
    \end{minipage}
    \vspace{-2mm}
    \caption{Time-wise entropy-increase agreement ratios computed from squared channel magnitudes.
    Each point is averaged over blocks and image tokens.}
    \label{fig:time-wise-agreement}
\end{figure}

\let\timetrendplot\relax

\subsection{Different resolution conclusion is stable}
\label{app:resolution}
We repeat the observation experiment in \cref{sec:analysis} at $1024\times1024$ resolution instead of the $512\times512$ resolution used in \cref{tab:feature-statistics}, using the same prompts, models, and detail-instruction suffixes.
\Cref{tab:feature-statistics-1024} reports the same statistics as \cref{tab:feature-statistics}.
The conclusion is unchanged: detail instructions increase the mean entropy of squared channel magnitudes and the entropy-agreement ratios more consistently than content instructions across resolutions.

\begin{table*}[t]
    \centering
    \resizebox{0.9\textwidth}{!}{%
        \renewcommand{\arraystretch}{1.0}
        \begin{tabular}{llccccc}
            \toprule
            Model & $\overline{H}(\rvp)$ & Added text & $\overline{H}(\hat{\rvp})$ & $A_{\mathrm{all}}$ & $A_{\mathrm{prompt}}$ & Spearman \\
            \midrule
            \multirow{2}{*}{FLUX.2 Klein 9B Base} & \multirow{2}{*}{4.510} & content & 4.507 & 48.7\% & 45.6\% & 0.967 \\
            & & \cellcolor{lightgreen}detail & \cellcolor{lightgreen}\textbf{4.561} & \cellcolor{lightgreen}\textbf{71.0\%} & \cellcolor{lightgreen}\textbf{100.0\%} & \cellcolor{lightgreen}\textbf{0.969} \\
            \midrule
            \multirow{2}{*}{FLUX.2 Klein 9B} & \multirow{2}{*}{4.355} & content & 4.328 & 38.8\% & 18.4\% & 0.846 \\
            & & \cellcolor{lightgreen}detail & \cellcolor{lightgreen}\textbf{4.442} & \cellcolor{lightgreen}\textbf{70.8\%} & \cellcolor{lightgreen}\textbf{100.0\%} & \cellcolor{lightgreen}\textbf{0.902} \\
            \midrule
            \multirow{2}{*}{SD3.5 Large} & \multirow{2}{*}{3.115} & content & 3.118 & 58.8\% & 64.0\% & 0.950 \\
            & & \cellcolor{lightgreen}detail & \cellcolor{lightgreen}\textbf{3.131} & \cellcolor{lightgreen}\textbf{69.8\%} & \cellcolor{lightgreen}\textbf{98.8\%} & \cellcolor{lightgreen}\textbf{0.971} \\
            \midrule
            \multirow{2}{*}{SD3.5 Large Turbo} & \multirow{2}{*}{3.013} & content & 3.020 & 57.5\% & 67.4\% & 0.908 \\
            & & \cellcolor{lightgreen}detail & \cellcolor{lightgreen}\textbf{3.037} & \cellcolor{lightgreen}\textbf{68.8\%} & \cellcolor{lightgreen}\textbf{98.0\%} & \cellcolor{lightgreen}\textbf{0.939} \\
            \midrule
            \multirow{2}{*}{Z-Image Base} & \multirow{2}{*}{0.055} & content & 0.056 & 51.5\% & 70.4\% & 0.869 \\
            & & \cellcolor{lightgreen}detail & \cellcolor{lightgreen}{0.056} & \cellcolor{lightgreen}{51.9\%} & \cellcolor{lightgreen}\textbf{84.4\%} & \cellcolor{lightgreen}\textbf{0.905} \\
            \midrule
            \multirow{2}{*}{Z-Image Turbo} & \multirow{2}{*}{1.083} & content & 1.083 & 56.5\% & 55.4\% & 0.888 \\
            & & \cellcolor{lightgreen}detail & \cellcolor{lightgreen}\textbf{1.090} & \cellcolor{lightgreen}\textbf{62.0\%} & \cellcolor{lightgreen}\textbf{92.8\%} & \cellcolor{lightgreen}\textbf{0.920} \\
            \bottomrule
        \end{tabular}%
    }
    \vspace{-1mm}
    \caption{Same comparison as \cref{tab:feature-statistics}, computed at $1024\times1024$ resolution instead of $512\times512$.
    The same pattern holds: detail instructions increase mean entropy and agreement ratios more than content instructions across models, confirming the conclusion is stable across resolutions.}
    \label{tab:feature-statistics-1024}
    \vspace{-2mm}
\end{table*}

\subsection{Distribution across channel-magnitude bins}
\label{app:channel-bin-shares}
We further examine how detail instructions redistribute the density (squared channel magnitudes) by computing their accumulated density.
We divide the ranked channel magnitudes into 64 equal-sized bins and average each bin's share over prompts, blocks, timesteps, and image tokens.
As an example, \cref{tab:top-bin-shares} reports the eight largest shares for the distilled FLUX.2 Klein 9B model.
With a detail instruction, the first bin contributes less, while the following bins contribute more, yielding a flatter distribution.

\begin{table}[t]
    \centering
    \renewcommand{\arraystretch}{1.05}
    \setlength{\tabcolsep}{5pt}
    \begin{tabular}{c@{\hspace{8pt}}cccccccc}
        \toprule
        \multirow{2}{*}{Detail instruction}
        & \multicolumn{8}{c}{Top eight shares among 64 bins} \\
        \cmidrule(l){2-9}
        & 1 & 2 & 3 & 4 & 5 & 6 & 7 & 8 \\
        \midrule
        & \textbf{0.6438} & 0.0416 & 0.0317 & 0.0262 & 0.0226 & 0.0198 & 0.0177 & 0.0159 \\
        $\checkmark$ & 0.6310 & \textbf{0.0430} & \textbf{0.0328} & \textbf{0.0272} & \textbf{0.0234} & \textbf{0.0205} & \textbf{0.0183} & \textbf{0.0165} \\
        \bottomrule
    \end{tabular}
    \caption{Top eight shares of the averaged 64-bin squared channel-magnitude distribution for the distilled FLUX.2 Klein 9B model, without and with a detail instruction.}
    \label{tab:top-bin-shares}
\end{table}

\section{More about the T$\flat$}

\subsection{A clarification about the token amplifier}
\label{app:token-amplifier-clarification}
For negative values of $\alpha$, the interpolation in \cref{eq:interpolate} may produce negative entries in $\rvp'$, 
violating the nonnegativity requirement of a density.
In practice, we apply ReLU to $\rvp'$ and then renormalize it to sum to one for the implementation.
For presentation simplicity, we omit these operations from the main formulation.

\subsection{Further implementation details.}
\label{app:implementation-details}
For FLUX.2 Klein 9B Base and its distilled variant, we inject T$\flat$ into all single-stream blocks (last 24 blocks).
Qwen-Image adopts a double-stream MMDiT architecture, and we apply T$\flat$ to its last 40 blocks.
For models distilled without CFG, we apply T$\flat$ only to the conditional branch.

For the comparison in \cref{tab:comparison}, we use the following method-specific settings:
T$\flat$ with $\alpha=+0.0045$; 
FDG with $s_{\mathrm{low}}=1.7$ and $s_{\mathrm{high}}=8$; 
HiGS with $s=1$; FreSca with $s=3$ and $r_0=0.5$ where $r_0$ is the low-frequency cutoff;
SEG with $s=3$; PAG with $s=1$; DG with $w=0.5$; TPG with $s=0.2$; SAG with $s=0.75$; and SSG with $s=0.2$.
$s$ means the guidance weight. 

\subsection{More about the competing method DG}
\label{app:dg-comparison}
Our findings share some similarities with DG~\citep{dg}: both study token-channel magnitudes, and both relate to image detail.
We clarify the differences below.

Mechanistically, the two guidance signals differ: 
DG requires an extra forward pass to construct its negative-guidance prediction, while T$\flat$ does not.
This is reflected in \cref{tab:comparison}, where DG costs $1.53\times$ the baseline latency versus $1.03\times$ for T$\flat$.

The underlying findings also differ.
DG observes massive activations: 
a few channels dominate a token, and masking them out causes only local distortion rather than structural change.
DG therefore uses the masked prediction as negative guidance.
This is effective at fixing local corruptions such as blur or unclear detail.
Their method primarily serves as a safeguard against local detail degradation rather than a mechanism for further enhancing already well-rendered details.
In contrast, we compare model behavior with and without a detail instruction and use that difference to construct a direction 
that increases detail continuously.

\subsection{More about time-aware effects.}
\label{app:time-aware-effects}
For the analysis in \cref{fig:time-step-energy}, we partition the radial Fourier spectrum into five bands:
$0$--$0.10\pi$, $0.10\pi$--$0.25\pi$, $0.25\pi$--$0.45\pi$, $0.45\pi$--$0.70\pi$, and $0.70\pi$--$\pi$.
Because natural images contain substantially more low-frequency than high-frequency energy, we report the relative energy change within each band rather than its absolute energy.
For the four-step FLUX.2 Klein 9B example analyzed in \cref{fig:time-step-energy,fig:time-step-lpips}, we place the scheduler peak between the second and third steps to balance structural preservation and high-frequency detail enhancement.

Would a low-level post-processing operation on the baseline, such as sharpening or boosting high frequencies, achieve the same effect?
This is essentially shown by the timestep comparison in \cref{fig:time-step-energy,fig:time-step-lpips}: applying such a low-level operation resembles pushing $\rho$ to a very late step.
However, low-level operations cannot generate new detail; they can only strengthen detail that is already present.
T$\flat$ can do both, and both regimes remain controllable through $\rho$ and $\kappa$.

\subsection{Hyperparameter adjustment}
\label{app:hyperparameter-adjustment}
The scheduler parameters $\rho$ and $\kappa$ in the experiments for different models are not fixed to a single choice.
As with the timestep analysis in \cref{fig:time-step-energy,fig:time-step-lpips}, 
moving the peak earlier or later can be a user choice: 
both directions are valid, 
and which one to pick depends on whether low-frequency structure or high-frequency detail should be emphasized.

We set $\rho=0.9$ for FLUX.2 Klein 9B because it is a distilled, 
four-step model whose final sampling step already sits at $\sigma_t\approx0.75$, 
so $\rho=0.9$ falls roughly between its first two steps.
FLUX.2 Klein 9B Base and Qwen-Image instead use $25$-step, 
non-distilled sampling, where image structure solidifies more slowly, 
so we push the peak slightly later, to $\rho=0.7$ and $\rho=0.8$ respectively.
In general, we find $\rho\in[0.85,0.95]$ works best for step-distilled models,
and $\rho\in[0.7,0.8]$ works best for non-distilled models.

\subsection{Performance of T$\flat$ on SD3.5 and Z-image families}
\label{app:sdc-families}
We extend our evaluation to the SD3.5 and Z-Image families, including both base and accelerated variants.
As discussed in \cref{sec:further-discussion}, the effectiveness of T$\flat$ depends on the capability of the pretrained model.
These models generally perform worse with T$\flat$ than the three models evaluated in the main paper.
We examine whether changing $\alpha$ still provides monotonic detail control and how it affects image quality and structural consistency.
The complete quantitative results are reported in \cref{tab:appendix-sdc-families}.

\begin{table}[t]
    \centering
    \resizebox{\textwidth}{!}{%
    \renewcommand{\arraystretch}{1.0}
    \begin{tabular}{l  cc  cccc  cc}
        \toprule
        \multirow{2}{*}{Input $\alpha$} & \multicolumn{2}{c}{Image Quality}
        & \multicolumn{4}{c}{Detail Statistics $\nearrow_\alpha$}
        & \multicolumn{2}{c}{Structure Difference} \\
        \cmidrule(lr){2-3}\cmidrule(lr){4-7}\cmidrule(lr){8-9}
        & CLIP$\uparrow$ & HPSv3$\uparrow$ & Edge$\uparrow$ & Sharpness$\uparrow$ & $f_{95}\uparrow$ & Wavelet$\uparrow$ & DINO & LPIPS$_{128}$ \\
        \midrule
        \rowcolor{modellavender}
        \multicolumn{9}{l}{\rule[-0.6ex]{0pt}{2.7ex}SD3.5 Large} \\
        $-0.03$ & 0.2722 & 11.2652 & 0.0317 & 0.0188 & 0.0377 & 0.0219 & 0.0933 & 0.0922 \\
        $-0.02$ & 0.2724 & 11.5866 & 0.0324 & 0.0193 & 0.0395 & 0.0232 & 0.0760 & 0.0767 \\
        $-0.01$ & 0.2726 & 11.8132 & 0.0331 & 0.0195 & 0.0416 & 0.0248 & 0.0600 & 0.0627 \\
        \rowcolor{lightgray}
        $0$ (baseline) & 0.2726 & \textbf{12.0507} & 0.0340 & 0.0194 & 0.0441 & 0.0268 & 0.0000 & 0.0000 \\
        $+0.01$ & 0.2727 & 11.9897 & 0.0351 & 0.0195 & 0.0461 & 0.0287 & 0.0607 & 0.0644 \\
        $+0.02$ & 0.2730 & 11.9015 & 0.0360 & 0.0194 & 0.0476 & 0.0301 & 0.0763 & 0.0784 \\
        $+0.03$ & 0.2724 & 11.4555 & 0.0369 & 0.0192 & 0.0490 & 0.0318 & 0.1002 & 0.1037 \\
        $\nearrow_{\alpha}$ & - & - & \textcolor{green!40!black}{$\checkmark$} & - & \textcolor{green!40!black}{$\checkmark$} & \textcolor{green!40!black}{$\checkmark$} & - & - \\
        \midrule
        \rowcolor{modelpeach}
        \multicolumn{9}{l}{\rule[-0.6ex]{0pt}{2.7ex}SD3.5 Large Turbo} \\
        $-0.15$ & 0.2676 & 10.6599 & 0.0232 & 0.0108 & 0.0182 & 0.0108 & 0.0897 & 0.0699 \\
        $-0.10$ & 0.2678 & 11.5693 & 0.0238 & 0.0111 & 0.0200 & 0.0120 & 0.0536 & 0.0475 \\
        $-0.05$ & 0.2678 & 11.8290 & 0.0242 & 0.0112 & 0.0208 & 0.0127 & 0.0329 & 0.0339 \\
        \rowcolor{lightgray}
        $0$ (baseline) & 0.2685 & \textbf{12.0571} & 0.0247 & 0.0112 & 0.0214 & 0.0134 & 0.0000 & 0.0000 \\
        $+0.05$ & 0.2689 & 11.9711 & 0.0249 & 0.0109 & 0.0214 & 0.0136 & 0.0311 & 0.0325 \\
        $+0.10$ & 0.2685 & 11.7018 & 0.0250 & 0.0107 & 0.0213 & 0.0136 & 0.0476 & 0.0458 \\
        $+0.15$ & 0.2679 & 11.1279 & 0.0252 & 0.0104 & 0.0213 & 0.0137 & 0.0799 & 0.0655 \\
        $\nearrow_{\alpha}$ & - & - & \textcolor{green!40!black}{$\checkmark$} & - & - & \textcolor{green!40!black}{$\checkmark$} & - & - \\
        \midrule
        \rowcolor{modelaqua}
        \multicolumn{9}{l}{\rule[-0.6ex]{0pt}{2.7ex}Z-Image Turbo} \\
        $-0.003$ & 0.2628 & 13.1598 & 0.0196 & 0.0075 & 0.0350 & 0.0171 & 0.0468 & 0.0385 \\
        $-0.002$ & 0.2630 & 13.1764 & 0.0198 & 0.0076 & 0.0355 & 0.0174 & 0.0301 & 0.0250 \\
        $-0.001$ & 0.2634 & \textbf{13.2295} & 0.0201 & 0.0079 & 0.0366 & 0.0180 & 0.0196 & 0.0148 \\
        \rowcolor{lightgray}
        $0$ (baseline) & 0.2641 & 13.1989 & 0.0206 & 0.0083 & 0.0382 & 0.0190 & 0.0000 & 0.0000 \\
        $+0.001$ & 0.2643 & 13.1321 & 0.0211 & 0.0086 & 0.0397 & 0.0201 & 0.0188 & 0.0136 \\
        $+0.002$ & 0.2647 & 12.9700 & 0.0216 & 0.0089 & 0.0412 & 0.0213 & 0.0397 & 0.0276 \\
        $+0.003$ & 0.2645 & 12.8665 & 0.0224 & 0.0092 & 0.0430 & 0.0230 & 0.0650 & 0.0460 \\
        $\nearrow_{\alpha}$ & - & - & \textcolor{green!40!black}{$\checkmark$} & \textcolor{green!40!black}{$\checkmark$} & \textcolor{green!40!black}{$\checkmark$} & \textcolor{green!40!black}{$\checkmark$} & - & - \\
        \midrule
        \rowcolor{modelsand}
        \multicolumn{9}{l}{\rule[-0.6ex]{0pt}{2.7ex}Z-Image Base} \\
        $-3 \times 10^{-5}$ & 0.2716 & 11.5622 & 0.0261 & 0.0181 & 0.0533 & 0.0284 & 0.0785 & 0.0724 \\
        $-2 \times 10^{-5}$ & 0.2719 & 11.6882 & 0.0270 & 0.0187 & 0.0549 & 0.0295 & 0.0675 & 0.0614 \\
        $-1 \times 10^{-5}$ & 0.2722 & \textbf{11.7037} & 0.0280 & 0.0191 & 0.0577 & 0.0311 & 0.0543 & 0.0467 \\
        \rowcolor{lightgray}
        $0$ (baseline) & 0.2725 & 11.6860 & 0.0290 & 0.0187 & 0.0602 & 0.0328 & 0.0000 & 0.0000 \\
        $+1 \times 10^{-5}$ & 0.2726 & 11.6299 & 0.0302 & 0.0183 & 0.0623 & 0.0347 & 0.0575 & 0.0538 \\
        $+2 \times 10^{-5}$ & 0.2729 & 11.3411 & 0.0313 & 0.0177 & 0.0631 & 0.0364 & 0.0869 & 0.0872 \\
        $+3 \times 10^{-5}$ & 0.2731 & 10.9888 & 0.0326 & 0.0176 & 0.0641 & 0.0386 & 0.1100 & 0.1286 \\
        $\nearrow_{\alpha}$ & - & - & \textcolor{green!40!black}{$\checkmark$} & - & \textcolor{green!40!black}{$\checkmark$} & \textcolor{green!40!black}{$\checkmark$} & - & - \\
        \bottomrule
    \end{tabular}%
    }
    \caption{T$\flat$ results for the SD3.5 and Z-Image families over a range of $\alpha$ values.}
    \label{tab:appendix-sdc-families}
    \vspace{-2mm}
\end{table}

\subsection{Detailed definitions of detail metrics}
\label{app:detail-metrics}
Let $I\in[0,1]^{H\times W\times 3}$ denote an RGB image at its native resolution.
All four detail metrics are computed from its Rec.709 luminance,
\begin{equation}
    Y_{ij}=0.2126R_{ij}+0.7152G_{ij}+0.0722B_{ij}.
\end{equation}

\paragraph{Edge strength.}
This metric measures the average strength of local intensity transitions and image boundaries.
We compute horizontal and vertical image gradients using central differences for interior pixels and one-sided differences at the boundary:
\begin{equation}
    (\partial_xY)_{ij}=\frac{Y_{i,j+1}-Y_{i,j-1}}{2},\qquad
    (\partial_yY)_{ij}=\frac{Y_{i+1,j}-Y_{i-1,j}}{2}.
\end{equation}
Edge strength is the mean gradient magnitude,
\begin{equation}
    E(I)=\frac{1}{HW}\sum_{i,j}
    \sqrt{(\partial_xY)_{ij}^{2}+(\partial_yY)_{ij}^{2}}.
\end{equation}
We use direct differences rather than Sobel filters to avoid attenuating the high-frequency components being measured.

\paragraph{Sharpness.}
This metric measures the strength of fine-scale intensity variation associated with image sharpness.
We apply the four-neighbor Laplacian to the $(H-2)(W-2)$ interior pixels,
\begin{equation}
    (\Delta Y)_{ij}=Y_{i-1,j}+Y_{i+1,j}+Y_{i,j-1}+Y_{i,j+1}-4Y_{ij},
\end{equation}
and define sharpness as its variance:
\begin{equation}
    S(I)=\operatorname{Var}_{i,j}\!\left[(\Delta Y)_{ij}\right].
\end{equation}

\paragraph{Spectral edge frequency.}
This metric measures at which frequency of the spectral domain, the cumulative energy extends 95$\%$ of the total cumulative energy.
To reduce boundary discontinuities in the Fourier transform, we first apply a separable Hann window
\begin{equation}
    w_{ij}=h_i^{(H)}h_j^{(W)},\qquad
    h_n^{(N)}=\frac{1}{2}\left(1-\cos\frac{2\pi n}{N-1}\right),
\end{equation}
and compute the centered power spectrum and normalized radial frequency:
\begin{align}
    P &= \left|\operatorname{fftshift}\mathcal{F}(Y\odot w)\right|^2,\\
    r_{ij} &= \sqrt{\left(\frac{i-H/2}{H/2}\right)^2+
    \left(\frac{j-W/2}{W/2}\right)^2}.
\end{align}
Let
\begin{equation}
    C(r)=\frac{\sum_{r_{ij}\le r}P_{ij}}{\sum_{i,j}P_{ij}}
\end{equation}
denote the cumulative fraction of spectral energy within radius $r$.
The spectral edge frequency is the smallest radius containing 95\% of the energy:
\begin{equation}
    f_{95}(I)=\inf\{r:C(r)\ge0.95\}.
\end{equation}
Normalizing each axis by its Nyquist frequency makes $f_{95}$ comparable across image resolutions and aspect ratios.

\paragraph{Wavelet detail energy.}
This metric measures the fraction of image energy contained in localized details across multiple spatial scales.
We apply a three-level orthonormal Haar transform to $Y$, starting from $A_0=Y$.
At each level, horizontal filtering gives
\begin{equation}
    u_{ij}=\frac{A_{i,2j}+A_{i,2j+1}}{\sqrt{2}},\qquad
    v_{ij}=\frac{A_{i,2j}-A_{i,2j+1}}{\sqrt{2}},
\end{equation}
followed by vertical filtering:
\begin{align}
    LL_{ij}&=\frac{u_{2i,j}+u_{2i+1,j}}{\sqrt{2}}, &
    LH_{ij}&=\frac{u_{2i,j}-u_{2i+1,j}}{\sqrt{2}},\\
    HL_{ij}&=\frac{v_{2i,j}+v_{2i+1,j}}{\sqrt{2}}, &
    HH_{ij}&=\frac{v_{2i,j}-v_{2i+1,j}}{\sqrt{2}}.
\end{align}
We recursively set $A_l=LL_l$ and measure the fraction of energy in all detail subbands:
\begin{equation}
    R_{\mathrm{wav}}(I)=
    \frac{\sum_{l=1}^{3}\left(\lVert LH_l\rVert_2^2+\lVert HL_l\rVert_2^2+\lVert HH_l\rVert_2^2\right)}
    {\lVert A_3\rVert_2^2+\sum_{l=1}^{3}\left(\lVert LH_l\rVert_2^2+\lVert HL_l\rVert_2^2+\lVert HH_l\rVert_2^2\right)}.
\end{equation}
Because the transform is orthonormal, the denominator equals the total luminance energy up to odd-size boundary cropping.

\subsection{More qualitative examples}
\label{app:qualitative}
Image quality assessment metrics may not fully reflect human preference,
and whether an image looks better is always subjective and biased.
Hence, we show some additional qualitative examples in this section.
We show some examples of T$\flat$ under different $\alpha$ selections and different models
in \cref{fig:additional-qualitative-examples-3,fig:additional-qualitative-examples-2,fig:additional-qualitative-examples-4,fig:additional-qualitative-examples-5,fig:additional-qualitative-examples}.
Instead of showing only the well-performing ones, we also show some representative failure cases.

One limitation of T$\flat$ is that it operates within a limited effective range.
An excessively large $\alpha$ may distort images, 
producing cartoon-like appearances, 
sharp artifacts, 
or unnatural background clutter.
The figures below provide additional qualitative examples to illustrate it more intuitively.
In addition, T$\flat$ may also shift the overall color tone of the images, not solely controlling the local details.

\input{figures/fig_add_example1.tex}

\input{figures/fig_add_example0.tex}

\input{figures/fig_add_example2.tex}

\input{figures/fig_add_example3.tex}

\input{figures/fig_additional_examples.tex}

\subsection{More comparisons with competing methods}
\label{app:more-comparisons}
\Cref{tab:comparison} reports quantitative results against the competing training-free detail-enhancement methods.
Here we provide additional qualitative comparisons, in the same format as \cref{fig:comparison}, but with two rows per prompt: the top row is unchanged (Baseline, T$\flat$, DG, FreSca, SAG), and the bottom row adds every remaining method except FDG (PAG, SEG, TPG, SSG, HiGS).

\input{figures/fig_add_compare0.tex}

\input{figures/fig_add_compare1.tex}

\input{figures/fig_add_compare2.tex}

\input{figures/fig_add_compare3.tex}

\input{figures/fig_add_compare4.tex}

\section{Future work}
\label{app:future-work}
\Cref{tab:feature-statistics} also shows that entropy does not increase for every token: 
the agreement ratio $A_{\mathrm{all}}$ stays well below 100\% for all models.
This suggests that each token likely has its own optimal direction for detail enhancement, rather than sharing a single one.
T$\flat$ instead pushes every token's channel density toward a more uniform distribution, which is unlikely to be optimal for every token.
However, characterizing a per-token optimal direction is difficult for a training-free method.
We hope this observation motivates future work on training a network that enables fine-grained, per-token control over detail.

\end{document}